\documentclass[a4paper,fleqn]{cas-sc}

\usepackage[numbers,sort&compress]{natbib}
\usepackage{hyperref}
\usepackage{graphicx}
\usepackage{subcaption}
\usepackage{booktabs}
\usepackage{multirow}
\usepackage{stfloats}
\usepackage{float}
\usepackage{lmodern}
\usepackage{setspace}

\usepackage{setspace}

\AtBeginDocument{
\renewcommand{\baselinestretch}{1.25}
\selectfont
}

\def\tsc#1{\csdef{#1}{\textsc{\lowercase{#1}}\xspace}}
\tsc{WGM}
\tsc{QE}
\tsc{EP}
\tsc{PMS}
\tsc{BEC}
\tsc{DE}

\begin{document}
\let\WriteBookmarks\relax
\def\floatpagepagefraction{1}
\shorttitle{TUSI: A Citation-Grounded Graph-Based Benchmark}
\shortauthors{Karimi et~al.}
\title [mode = title]{TUSI: A Citation-Grounded Graph-Based Benchmark for Retrieval-Augmented Generation in Persian Legal Question Answering}


\author[1]{Marzieh Karimi}
\cormark[1]
\ead{mkarimi@ec.iut.ac.ir}
\credit{Conceptualization, Methodology, Software, Data curation, 
Writing -- original draft, Writing -- review and editing}

\author[1]{Mehran Safayani}
\credit{Methodology, Data curation, Writing -- review and editing}
\ead{safayani@iut.ac.ir
}
\author[1]{Alireza Basiri}
\credit{Methodology, Validation, Writing -- review and editing}
\ead{basiri@iut.ac.ir}

\affiliation[1]{organization={Department of Electrical and Computer Engineering, Isfahan University of Technology},
                city={Isfahan},
                postcode={84156-83111},
                country={Iran}}

\begin{abstract}
Reliable evaluation of retrieval-augmented generation (RAG) systems in the legal domain remains challenging, particularly for low-resource languages such as Persian, where available benchmarks are limited in realism, grounding, and coverage of legal reasoning. Existing resources are either synthetically generated from statutory texts, lacking natural information needs, or derived from informal legal question--answer sources without explicit linkage to authoritative legal documents, making them unsuitable for rigorous evaluation of grounded legal reasoning.
To address this gap, we introduce {TUSI-Bench}, a structured Persian legal benchmark designed for the evaluation of RAG-based systems. TUSI-Bench is built upon a multi-layered legal knowledge framework containing 7,261 legal articles and integrating codified statutes, legal commentaries, doctrinal writings, and Supreme Court unification rulings. In addition, we construct a legal knowledge graph that explicitly models relationships among legal articles and concepts to support structure-aware retrieval.
Based on this foundation, TUSI-Bench provides two complementary evaluation tasks: (i) {TUSI-QA}, an open-ended question answering benchmark containing 1,999 questions that requires grounded responses with explicit legal citations, and (ii) {TUSI-Bar}, a multiple-choice benchmark containing approximately 500 four-option questions derived from Iranian Bar examination questions. Furthermore, we propose a graph-based retrieval baseline that leverages relational structure within the legal knowledge graph.
Experimental results demonstrate that incorporating graph-based relations improves retrieval recall by up to 8 percentage points compared to standard baselines, highlighting the importance of structured legal representations for RAG systems. The proposed benchmark provides a standardized resource for evaluating grounded legal reasoning in Persian and enables future research on structured legal intelligence.
\end{abstract}


\begin{highlights}
\item TUSI-Bench enables grounded RAG evaluation in Persian legal QA.
\item TUSI-QA and TUSI-Bar cover complementary legal reasoning tasks.
\item A legal knowledge graph supports structure-aware retrieval.
\item Graph-based retrieval improves recall by up to 8 percentage points.
\end{highlights}


\maketitle

\section{Introduction}

Recent advances in artificial intelligence have led to the widespread adoption of large language models (LLMs) across diverse domains, including education, healthcare, and particularly the legal domain \cite{bommasani2021foundation}. In legal applications, these models are increasingly used to support tasks such as legal information retrieval, question answering, and case analysis \cite{li2025cole,zheng2025legal}. The appeal is clear: legal knowledge is expressed in dense natural-language text, and access to professional legal advice is costly and unevenly distributed. However, the reliability of such systems remains a critical concern in a high-stakes domain such as law, where incorrect or unsupported outputs may lead to serious legal and ethical consequences.

Working with real legal data exposes several well-documented weaknesses of LLM-based systems. First, LLMs are prone to hallucination, i.e., generating fluent but factually unsupported content \cite{ji2023survey}. In the legal domain this takes a particularly harmful form: recent studies show that general-purpose models frequently fabricate case law and misstate legal holdings \cite{dahl2024large}, and that even commercial legal research tools built on retrieval-augmented pipelines still produce incorrect or misgrounded answers at a non-trivial rate \cite{magesh2025hallucination}.

Second, legal texts are long, densely cross-referenced, and written in a formal register with complex syntax and specialized terminology, so that even foundational tasks such as named entity recognition require domain-specific adaptation \cite{KEY_NER}; this register also differs markedly from the everyday language in which laypeople describe their problems, and as a result the lexical and semantic gap between a user's narrative question and the governing provision is large, which makes retrieval difficult \cite{zheng2025legal,louis2024interpretable}. Third, a legally sound answer rarely depends on a single passage: it often requires combining several articles, their authoritative interpretations, and judicial rulings, whereas LLMs tend to under-use evidence placed in long contexts \cite{liu2024lost}. Fourth, legal knowledge is jurisdiction-specific and changes over time, whereas the knowledge stored in model parameters is static, and may reflect a different jurisdiction or an outdated version of the law. Finally, evaluating legal outputs is intrinsically hard: legal tasks demand multi-step reasoning that goes beyond surface-level text understanding \cite{guha2023legalbench}, and open-ended answers admit multiple acceptable formulations, so reference-based metrics alone are insufficient.

Legal systems worldwide are broadly classified into two major traditions: civil law (Romano-Germanic) systems and common law systems. In civil law systems---including the Iranian legal system, which is the focus of this study---sound legal reasoning requires strict adherence to codified statutory texts, legal doctrines, and authoritative sources, in contrast to common law systems, where judicial precedent plays the primary role. This dependence on explicit, codified legal evidence highlights a fundamental limitation of current LLMs: their knowledge is implicitly encoded within static parameters and cannot naturally adapt to continuously evolving legal systems. Consequently, ensuring factual grounding, interpretability, and up-to-date legal reasoning remains a persistent challenge, and it is particularly acute in codified legal systems such as Iran's, where a legally sound answer must be explicitly traceable to a specific statutory article.

To address these limitations, Retrieval-Augmented Generation (RAG) \cite{lewis2020rag} has emerged as a promising approach. By integrating external knowledge sources at inference time, RAG-based methods reduce reliance on parametric memory, enable dynamic retrieval of relevant legal documents, and make it possible to attach explicit citations to generated answers. Nevertheless, the development of robust legal RAG systems critically depends on high-quality evaluation benchmarks that can jointly assess retrieval effectiveness and grounded generation in domain-specific settings, since errors in either component propagate to the final answer \cite{pipitone2024legalbenchrag}.

Several English-language legal benchmarks have been introduced in this direction. Among the most notable are the Competition on Legal Information Extraction and Entailment (COLIEE) \cite{goebel2024coliee} and CLERC \cite{hou2025clerc}, which primarily focus on legal retrieval and entailment, as well as reasoning-oriented resources such as the Reasoning-Focused Legal Retrieval Benchmark \cite{zheng2025legal}, which introduces more complex legal question answering grounded in document retrieval. Broader suites such as LegalBench \cite{guha2023legalbench} cover a wide range of legal reasoning tasks, but they are built on English-language legal materials and do not cover Persian or the Iranian legal system.

These English-centric resources cannot simply be reused for Persian legal applications, for three main reasons. First, {law is jurisdiction-bound}. Substantive rules, legal concepts, and procedural requirements differ across countries, so a model or benchmark grounded in U.S., European, or other legal systems cannot validate answers about Iranian law; an answer that is correct under one jurisdiction may be wrong, or even meaningless, under another. The Iranian legal system is a codified civil-law system whose sources, structure, and doctrinal foundations differ substantially from those of common-law jurisdictions, and it includes judicial institutions specific to its system, such as the binding Unification Rulings of the General Assembly of the Supreme Court. Second, {the structure of legal evidence differs}: many English benchmarks are case-based, with queries and relevant documents linked through judicial citations, and those that are statute-based cover other jurisdictions, whereas Iranian legal reasoning is anchored in individually numbered statutory articles, so retrieval must be evaluated at the article level. Third, {translation does not bridge the gap}: translating English benchmarks into Persian would change the language but not the underlying law, and it would also erase the terminology, register, and phrasing of real Persian legal consultations.

The Persian language adds a further layer of difficulty. Persian is a comparatively low-resource language in NLP: annotated corpora, retrieval benchmarks, and domain-specific resources are scarce relative to English, and multilingual models are trained on far less Persian text. Persian is written in a right-to-left Perso-Arabic script with frequent orthographic variation (e.g., Arabic versus Persian character forms, and inconsistent use of the zero-width non-joiner), which complicates normalization, matching, and indexing. Persian legal language is moreover a specialized register that mixes modern Persian with Arabic-derived legal terminology and archaic constructions, and its vocabulary is rarely seen in general-purpose training data. In addition, tokenizers trained on predominantly English data tend to fragment non-English text into more tokens, which increases cost and shortens the effective context available to the model \cite{petrov2023language}, a significant handicap when legal provisions and their interpretations are long. As a result, both retrieval models and generators tend to perform less reliably on Persian legal text.

Despite these needs, existing resources for Persian legal NLP remain very limited. Available datasets are confined to a few repositories on HuggingFace \cite{iran_legal_persian_qa,iran_madani_law_dataset} and generally fall into two categories: either synthetically generated question sets derived directly from civil law articles---lacking realistic information needs and natural query complexity---or crawled question--answer pairs from legal forums, where answers provided by legal practitioners are not systematically linked to legal articles and are neither curated nor structured for evaluation purposes. General-purpose Persian QA datasets, in turn, are mostly extractive and non-legal. As a result, no existing Persian resource provides a reliable benchmark for rigorous evaluation of retrieval-augmented legal systems.

To address this gap, we introduce {TUSI-Bench}, a structured Persian legal evaluation benchmark specifically designed for assessing retrieval-augmented legal systems. The name {TUSI} is inspired by Nasir al-Din al-Tusi, the renowned Persian polymath and scholar. TUSI-Bench is built on a multi-layer legal knowledge framework of 7,261 statutory articles, linked to expert interpretations and Supreme Court Unification Rulings through a legal knowledge graph. On top of this framework, it provides two complementary components: {TUSI-QA}, an open-ended question answering benchmark of 1,999 real lawyer--client questions with explicit, validated legal citations, and {TUSI-Bar}, a four-option multiple-choice benchmark derived from Iranian Bar examination questions. We further propose a structure-aware retrieval baseline that exploits the relational structure of the knowledge graph.

Our main contributions are summarized as follows:
\begin{itemize}
\item We introduce {TUSI-Bench}, to the best of our knowledge among the first structured Persian legal benchmarks designed for evaluating RAG systems, addressing both the scarcity of Persian legal resources and the jurisdictional mismatch of English-centric benchmarks. It comprises two complementary components: {TUSI-QA}, covering citation-grounded open-ended legal question answering, and {TUSI-Bar}, covering four-option multiple-choice legal reasoning.
\item We construct a legal knowledge graph that links statutory articles, expert interpretations, and Unification Rulings, and use it to define a structure-aware retrieval baseline that improves Recall@5 by up to 8 percentage points over its dense-retrieval counterpart.
\item Unlike existing Persian legal datasets, TUSI-Bench connects statutory texts with naturally formulated legal questions and evidence-grounded answers, validated through an iterative agent-and-expert filtering process, providing a realistic and challenging evaluation setting.
\end{itemize}

The remainder of this paper is organized as follows.
Section~\ref{sec:related_work} reviews related work in legal NLP and retrieval-augmented systems. Section~\ref{sec:knowledge_layer} describes the dataset construction process and data sources. Section~\ref{sec:benchmark} presents the structure of the proposed benchmarks and annotation methodology. Section~\ref{sec:results} reports the experimental setup, evaluation results, and discussion. Finally, Section~\ref{sec:conclusion} concludes the paper and outlines future research directions.

\section{Related Work}
\label{sec:related_work}

This section situates TUSI-Bench with respect to four lines of prior work: general-domain retrieval and question answering (QA) benchmarks, legal retrieval and QA benchmarks, graph-based retrieval-augmented generation (RAG) for legal text, and Persian-language benchmarks. For each, we state what existing resources provide and which requirements of Persian statutory legal QA they leave unaddressed.

\subsection{General-Domain Retrieval and Question Answering}
\label{sec:rw_general}

Open-domain QA benchmarks such as Natural Questions (NQ)~\cite{kwiatkowski2019natural} and HotpotQA~\cite{yang2018hotpotqa} evaluate systems that retrieve evidence from Wikipedia and answer questions over it, and they are widely used to evaluate RAG~\cite{lewis2020rag}. BEIR~\cite{thakur2021beir} unifies 18 retrieval datasets across 9 task types. In these benchmarks, answers are typically short spans (or yes/no), and relevant passages share considerable vocabulary with the query; Zheng et al.~\cite{zheng2025legal} show that the query--passage lexical similarity of NQ and HotpotQA is markedly higher than in reasoning-intensive legal retrieval. HotpotQA adds multi-hop reasoning across several documents, but its evidence remains unstructured text rather than a structured body of rules with cross-references.

Recent retrieval benchmarks reduce this reliance on surface matching. BRIGHT~\cite{su2025bright} collects realistic reasoning-intensive queries in 12 datasets from domains such as economics, psychology, mathematics, and coding, and BIRCO~\cite{wang2024birco} evaluates retrieval under multi-faceted objectives. Neither targets legal text, so neither captures the structure of statutory provisions, their cross-references, or the need to ground an answer in specific articles.

\subsection{Legal Retrieval and Question Answering Benchmarks}
\label{sec:rw_legal}

Legal text differs from general-domain text in its formal register, its dense cross-referencing, and the requirement that conclusions be traceable to authoritative sources.

\paragraph{Case-law retrieval.}
Many legal retrieval datasets are built from citation links between court decisions. CLERC~\cite{hou2025clerc} derives queries from passages of U.S.\ federal case law and asks systems to retrieve the precedents cited in them; it also includes a task of generating legal analysis from retrieved cases, in which models frequently hallucinate citations. The case-law tasks of COLIEE~\cite{goebel2024coliee} follow a similar query-case-to-supporting-case setup over Canadian Federal Court decisions, with entire case documents serving as queries (thousands of tokens on average, Table~\ref{tab:full_comparison}). Because cited rules are often restated in the citing text, such queries tend to share much vocabulary with their relevant documents~\cite{zheng2025legal}. These resources therefore differ from ours in two ways: their queries are long judicial texts rather than short natural-language questions from people seeking legal advice, and their evidence is unstructured precedent text rather than numbered statutory articles.

\paragraph{Statute retrieval and question answering.}
The statute-law tasks of COLIEE~\cite{goebel2024coliee} are the closest precedent to our setting: given a yes/no question from the Japanese bar exam, systems retrieve relevant Civil Code articles (Task~3) and decide whether they entail the answer (Task~4). These tasks cover a single legal system and use short, hypothetical exam questions with binary answers, so they cannot assess whether a system can explain, interpret, and cite the supporting provisions in a free-text answer to a real consultation. The reasoning-focused benchmark of Zheng et al.~\cite{zheng2025legal} comprises Bar Exam QA (bar exam fact patterns paired with judicial-opinion passages) and Housing Statute QA (6,853 questions over U.S.\ state statutes, with a corpus of about 1.8 million passages). It shows that low lexical overlap between queries and relevant text defeats lexical retrievers such as BM25, a property that TUSI-QA shares and even exceeds (Table~\ref{tab:lexical_similarity}). It is, however, built entirely on English-language U.S.\ materials and therefore cannot validate answers about Iranian law.

\paragraph{Broader legal suites.}
LegalBench~\cite{guha2023legalbench} is a collaboratively built benchmark of 162 tasks covering six types of legal reasoning (issue-spotting, rule-recall, rule-application, rule-conclusion, interpretation, and rhetorical understanding). Because the relevant legal text is supplied in the prompt, it evaluates reasoning but not retrieval. LegalBench-RAG~\cite{pipitone2024legalbenchrag} isolates the retrieval step of legal RAG, with 6{,}858 query--snippet pairs over a corpus of about 79M characters; however, it is built from contract and privacy-policy tasks, evaluates span-level rather than article-level retrieval, and does not evaluate answer generation. MSLR~\cite{yu2026mslr} evaluates multi-step legal reasoning over 1{,}389 Chinese insider-trading judgments, with about 60{,}000 annotated step-level reasoning paths based on the IRAC (Issue--Rule--Application--Conclusion) framework. It shows that current models remain limited in multi-step legal reasoning, but it targets reasoning quality within a single offense category of the Chinese legal system rather than statute-grounded retrieval and citation.

Overall, these benchmarks cover a handful of jurisdictions and languages (English/U.S., Japanese, and Chinese), and none combines a statutory corpus with both open-ended, citation-grounded QA and exam-style multiple-choice evaluation.

\subsection{Graph-Based Retrieval-Augmented Generation for Legal Text}
\label{sec:rw_graphrag}

Legal evidence is rarely confined to a single passage: a sound answer may depend on an article, the provisions it refers to, expert interpretations, and judicial rulings. Conventional RAG treats a corpus as a flat collection of passages and cannot follow such relations. Graph-based RAG (GraphRAG) addresses this by organizing the corpus as a graph and using its structure during retrieval and generation. In the original GraphRAG~\cite{edge2024graphrag}, an LLM extracts entities and relations from the corpus into a knowledge graph, the graph is partitioned into communities, and answers to corpus-level questions are built from community summaries, improving the comprehensiveness and diversity of the answers. Surveys describe a common workflow of graph-based indexing, graph-guided retrieval, and graph-enhanced generation~\cite{peng2024graphrag}.

Structure-aware retrieval has also been explored for statutes. For statutory article retrieval, Louis et al.~\cite{louis2023finding} derive a graph of the legislative structure from the division headings of the Belgian codes and apply graph neural networks to it, reporting that the hierarchy of the law is a useful retrieval signal. More recent work adapts GraphRAG to legal corpora explicitly. LegalGraphRAG~\cite{chen2026legalgraphrag} argues that legal corpora are heterogeneous, mixing cases, articles, and interpretations, and that a flat graph cannot separate them. It organizes legal knowledge into a hierarchical graph with ontology, fact, and rule layers and reasons over it with three cooperating agents (researcher, auditor, and adjudicator), and it is applied to criminal charge prediction. Oliveira et al.~\cite{oliveira2026propor} model Portuguese normative documents as graphs whose nodes are structural units (documents, articles, paragraphs, and items) and whose edges are normative relationships; their retriever assembles evidence along root-to-leaf paths before generation, and they evaluate it with RAGAS on 203 internal resolutions of a single energy company.

These works propose methods for individual corpora: the Portuguese study is limited to the internal resolutions of one organization, and LegalGraphRAG targets criminal charge prediction rather than statute-grounded question answering. We are not aware of any publicly available benchmark for Persian law that pairs an article-level legal knowledge graph with questions whose supporting provisions are annotated.

\subsection{Persian Benchmarks}
\label{sec:rw_persian}

Persian is a comparatively low-resource language in NLP, and the available benchmarks are dominated by general-domain tasks.

\paragraph{General question answering and evaluation.}
ParSQuAD~\cite{Abadani2021ParSQuAD}, PQuAD~\cite{darvishi2023pquad}, and PersianQuAD~\cite{persianquad2022} follow an extractive reading-comprehension format and are commonly used for Persian RAG-style evaluation; they assess span extraction rather than reasoning over legal statutes. Multiple-choice benchmarks evaluate general model ability. The Khayyam Challenge (PersianMMLU)~\cite{khayyam2024} contains 20{,}192 four-choice questions from 38 subjects of Persian school examinations. PerMMLU~\cite{permmlu2025} is a localized and extended Persian counterpart of MMLU that covers school-level, general, and specialized knowledge as well as knowledge specific to Persian-speaking societies; its general-knowledge subset touches on topics such as national laws, but only as isolated questions. PerCoR~\cite{alikhani2025percor} offers 106K multiple-choice sentence-completion problems for commonsense reasoning. None of these benchmarks includes a statutory corpus or involves retrieval.

\paragraph{Persian legal resources.}
Publicly released resources for Persian law are scarce. To our knowledge, the only openly available Persian legal QA collections are a few repositories on Hugging Face, such as the Iran Legal FAQ datasets~\cite{iran_legal_persian_qa,iran_madani_law_dataset}. They consist either of synthetically generated questions derived directly from legal articles, which lack realistic information needs, or of question--answer pairs crawled from legal forums, whose answers are explanatory, context-dependent, and not curated for evaluation. Crucially, neither kind provides ground-truth labels linking each question to the statutory articles that support its answer. To the best of our knowledge, no publicly released Persian legal dataset combines real-world legal questions with verified, article-level ground-truth labels, and as a result there is currently no way to evaluate whether a Persian legal system retrieves and cites the correct provisions.

To the best of our knowledge, no Persian benchmark combines
\begin{enumerate}
    \item a corpus of statutory legal texts,
    \item real-world legal questions,
    \item bar exam-style multiple-choice questions,
    \item reasoning-intensive retrieval with low lexical overlap between queries and relevant laws, and
    \item an article-level legal knowledge graph with annotated supporting provisions.
\end{enumerate}
TUSI-Bench addresses this gap by providing a Persian legal benchmark grounded in statutory law, designed to evaluate retrieval-based legal reasoning under realistic exam and practice settings.

\begin{figure}
\centering
\includegraphics[width=1\textwidth]{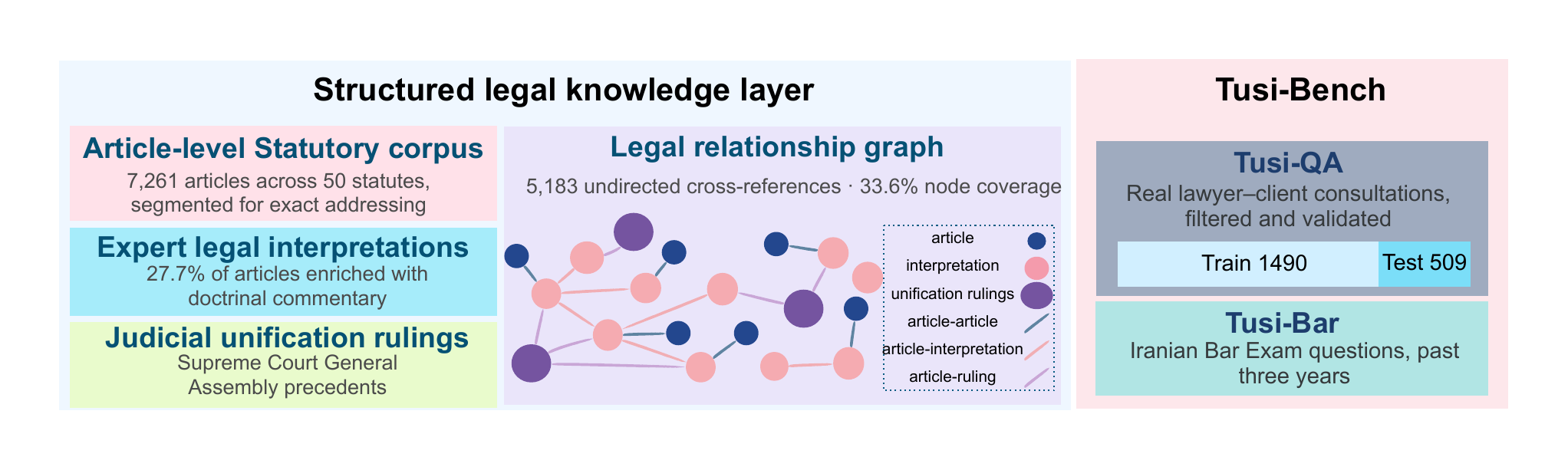}
\caption{Overview of the structured legal knowledge layer and the proposed {TUSI-Bench}. {TUSI-Bench} consists of two complementary subsets: {TUSI-QA}, a citation-grounded question answering set derived from real lawyer--client consultations; and {TUSI-Bar}, a multiple-choice question set derived from Iranian Bar Exam questions.}

\label{fig:knowledge_architecture}
\end{figure}

\section{Structured Legal Knowledge Layer}
\label{sec:knowledge_layer}
Our dataset is designed to support two primary tasks: Legal Information Retrieval and Legal Question Answering. In this setting, a system is expected to retrieve the relevant legal provisions for a given query and generate an answer grounded in statutory law and authoritative legal sources. Accordingly, the dataset is not limited to isolated question–answer pairs; instead, it incorporates a structured legal knowledge layer to enable systematic evaluation of retrieval-based and evidence-grounded legal systems. 
An overview of the resulting dataset architecture is presented in Figure~\ref{fig:knowledge_architecture}. 
The framework integrates a structured legal knowledge layer with benchmark-oriented evaluation resources, forming a unified foundation for legal information retrieval and citation-grounded question answering.

The dataset is organized as a multi-layer legal knowledge architecture consisting of: (1) statutory laws and executive regulations, (2) Judicial Unification Rulings, (3) expert interpretations and legal doctrine, and (4) a graph of inter-article relationships. This structure enables evaluation not only of surface-level retrieval, but also of multi-hop reasoning, cross-source retrieval, and interpretation-aware legal question answering. The outcome of this stage is a segmented statutory corpus accompanied by a legal relationship graph and complementary interpretative and judicial layers.

\subsection{Article-Level Statutory Corpus}

To construct the core legal knowledge layer, we selected a set of principal Iranian statutes that serve as foundational sources in formal legal education and the national bar examination. The selection criteria included: (i) frequency of citation in lawyers’ responses within the raw QA corpus, and (ii) recognized doctrinal and examination authority. The full texts of these statutes, relevant executive regulations, and Judicial Unification Rulings were collected from official and authoritative sources.

A preprocessing pipeline was applied, including removal of non-textual elements, orthographic normalization, and standardization of spacing and punctuation. For fine-grained retrieval and precise referencing, statutory texts were segmented at the level of articles. Each segment was assigned a unique identifier to enable exact addressing during retrieval, evaluation, and QA alignment. This design allows performance assessment not only at the statute level but also at the article level, which is essential for rigorous Legal Information Retrieval evaluation.

To enrich the semantic depth of the corpus, a complementary layer of expert legal interpretation was incorporated. These interpretations were extracted from a specialized legal knowledge source, where legal experts provide doctrinal explanations of complex statutory provisions grounded in authoritative legal scholarship and interpretative practice~\cite{edalatsara}. Each interpretative unit was explicitly linked to its corresponding statutory article. This enrichment supports interpretation-aware retrieval and retrieval-augmented legal reasoning.
An example of a statutory article and its corresponding expert interpretation is presented in Figure~\ref{fig:article10_structure}.

\subsection{Interpretation-Enriched Knowledge Graph}

Beyond statutory text and interpretation, inter-article relationships were systematically extracted in two ways. First, if one article explicitly referenced another within its text, we established a direct link between those two articles. Second, in the expert interpretations, when a connection to a particular statutory article was mentioned, a link was created between the interpretation and the referenced article.

Based on these references, we constructed a legal relationship graph. Formally, the legal knowledge graph is defined as $G = (V, E)$, where:

\begin{itemize}
    \item $V$ consists of nodes representing:
    \begin{itemize}
        \item statutory articles,
        \item interpretative units, and
        \item Unification Rulings, examples of which are shown in Figure~\ref{fig:unification_judgments_examples};
    \end{itemize}
    \item $E$ consists of typed edges representing:
    \begin{itemize}
        \item article-article references,
        \item article-interpretation links, and
        \item article-ruling associations.
    \end{itemize}
\end{itemize}

This graph structure enables the evaluation of graph-aware retrieval methods and multi-hop reasoning scenarios, particularly in cases where a legally sound answer requires combining multiple interrelated provisions. By incorporating these structured dependencies, the dataset supports experimentation with graph-based ranking, multi-source retrieval, and structured legal inference.

\subsection{Corpus Statistics}

\begin{figure}
\centering
\includegraphics[width=0.95\textwidth]{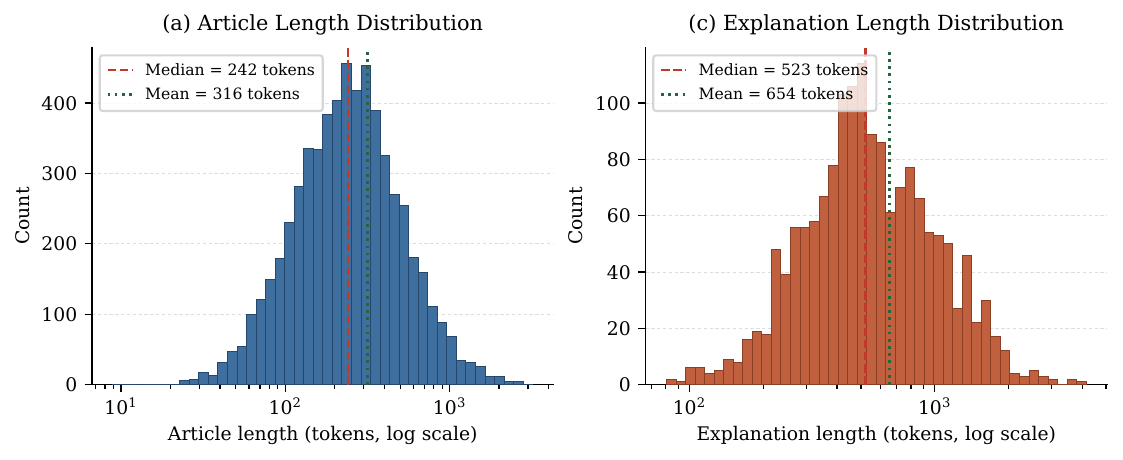}
\caption{Length distributions of statutory articles (left) and their corresponding legal interpretations (right).}
\label{fig:dis_art_interp_len}
\end{figure}

\begin{figure}
    \centering
    \includegraphics[width=0.6\columnwidth]{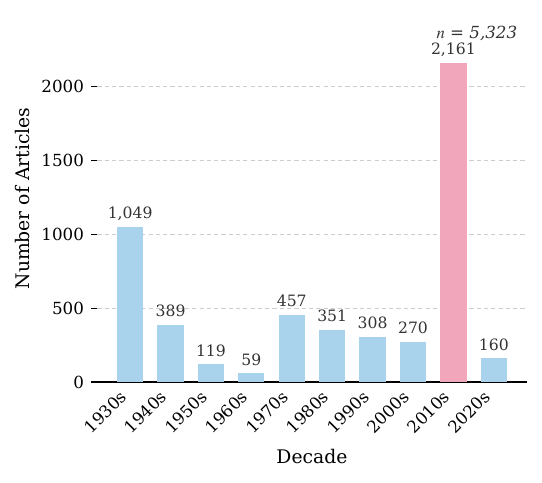}
    \caption{Temporal distribution of legal articles in the dataset across decades. The total number of articles is (n = 5{,}323).}
    \label{fig:temporal_distrib}
\end{figure}

\begin{figure}
\centering
\includegraphics[width=1\textwidth]{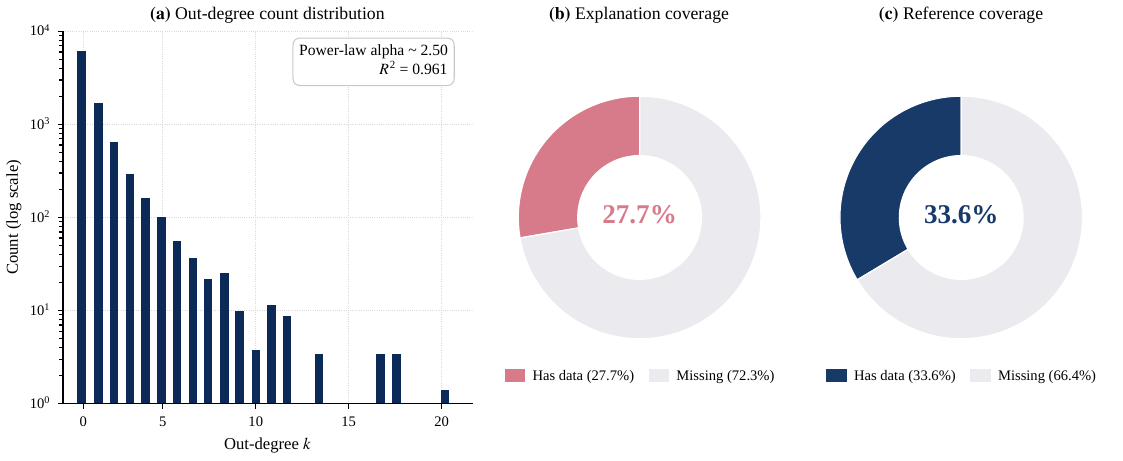}
\caption{Structural properties of the legal reference graph. (a) Out-degree distribution on a log scale, showing a heavy-tailed pattern consistent with a power-law ($\alpha = 2.50$, $R^2 = 0.961$), indicating highly uneven citation behavior. (b) Coverage of explanation annotations across nodes, where 27.7\% of articles have associated interpretations. (c) Coverage of reference annotations, showing that 33.6\% of nodes participate in at least one reference link, highlighting moderate but sparse connectivity in the graph structure.}
\label{fig:ref_graph}
\end{figure}

\begin{figure}
\centering
\includegraphics[width=1\textwidth]{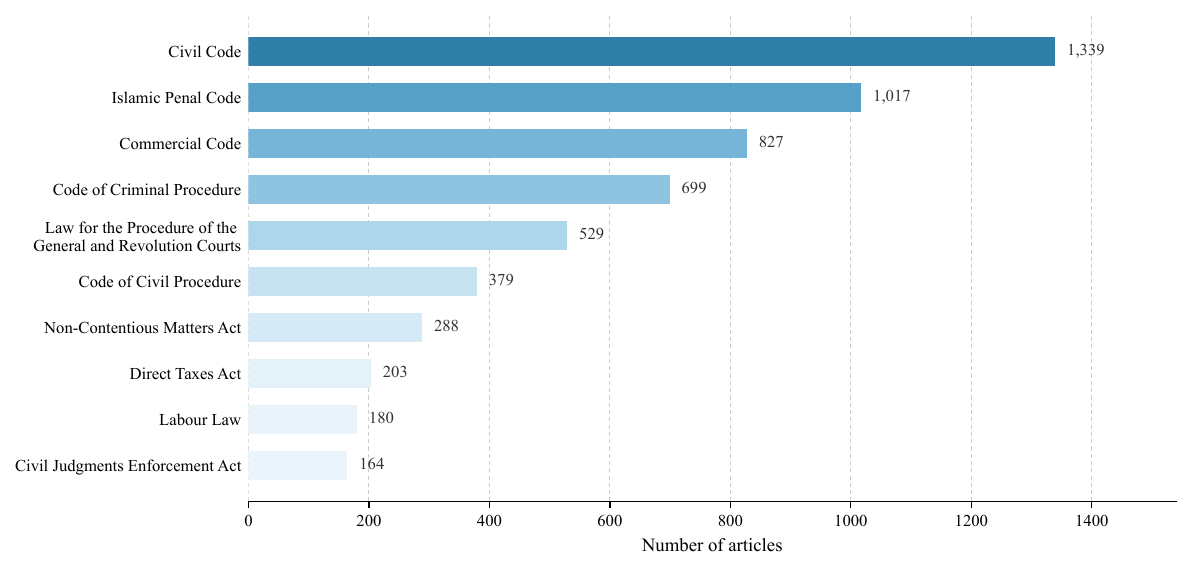}
\caption{Top 10 laws by article count in the dataset.}
\label{fig:top_laws}
\end{figure}

To characterize the structural, textual, and temporal properties of the legal corpus, we conduct a comprehensive quantitative analysis covering document length distributions, graph connectivity, coverage patterns, and temporal dynamics (Figures~\ref{fig:dis_art_interp_len}, \ref{fig:temporal_distrib}, \ref{fig:ref_graph} and \ref{fig:top_laws}). The dataset comprises 50 statutes and 7,261 legal articles, forming a heterogeneous legal knowledge graph with 9,272 nodes and 5,183 undirected cross-references.

The corpus exhibits strong textual heterogeneity with highly skewed, heavy-tailed length distributions. As illustrated in Figure~\ref{fig:dis_art_interp_len}, article lengths show a pronounced imbalance, with a median of 240 tokens and a mean of 370 tokens, indicating the presence of a large number of short provisions alongside a smaller set of substantially longer legal texts. In contrast, legal interpretations (when available) are consistently longer, with a median of 538 tokens and a mean of 695 tokens, reflecting their role as expert-driven expansions of selected statutory provisions. Importantly, these interpretations are not uniformly required across all articles; rather, they are selectively added by domain experts to clarify complex or ambiguous provisions, while many articles remain semantically self-contained.

The interpretative layer covers 27.7\% of articles (Figure~\ref{fig:ref_graph}), which should not be interpreted as a deficiency in the corpus, but rather as an intentional and selective enrichment process targeting legally or semantically complex cases. This results in a partially augmented knowledge structure where only a subset of provisions is accompanied by explanatory commentary.

From a structural perspective, the reference network is highly sparse and exhibits a clear power-law degree distribution ($\alpha = 2.50$, $R^2 = 0.961$), as shown in Figure~\ref{fig:ref_graph}. A large proportion of nodes (66.4\%) have zero out-degree, while a small number of highly connected nodes act as structural hubs, indicating an uneven distribution of cross-referencing activity across legal provisions.

The corpus also demonstrates strong temporal heterogeneity (Figure~\ref{fig:temporal_distrib}), spanning multiple decades from early legislative records to contemporary statutes, with a noticeable concentration in recent decades, particularly the 2010s. This reflects both the historical depth and the modern expansion of the legal system represented in the dataset.

Finally, the distribution of articles across statutes is highly imbalanced (Figure~\ref{fig:top_laws}), with a small number of laws accounting for a large fraction of the corpus. This long-tail pattern further reinforces the non-uniform nature of the dataset at both the document and structural levels.

By capturing the full and realistic diversity of Iranian legal rules across textual, structural, and temporal dimensions, this corpus forms a highly complex and heterogeneous legal knowledge graph, which in turn makes tasks such as information retrieval, reasoning, and graph-based learning particularly challenging.

\section{TUSI-Bench: A Citation-Grounded Legal QA Benchmark}
\label{sec:benchmark}
This section details the process of creating {TUSI-Bench} ({Persian LEgal eXam \& consultation Benchmark}), a hybrid question-answer dataset tailored for the Persian legal domain. The objective of this benchmark is to provide a comprehensive resource for the simultaneous evaluation of legal document retrieval systems and legal question-answering models. TUSI-Bench consists of two complementary components: (1) {TUSI-QA}, a citation-grounded Question-Answer set extracted from real user-lawyer interactions, and (2) {TUSI-Bar}, a Multiple-Choice Question set extracted from formal legal exams.

TUSI-QA includes real user questions answered by legal professionals and is based on two core principles: first, the use of actual data derived from real user-lawyer interactions; and second, ensuring that the answers are directly traceable to specific legal texts. TUSI-Bar is derived from the Iranian bar exam questions of the past three years.

\subsection{TUSI-QA: Consultation-Driven QA with Legal Citations}

\begin{figure}
    \centering
    \includegraphics[width=0.7\textwidth]{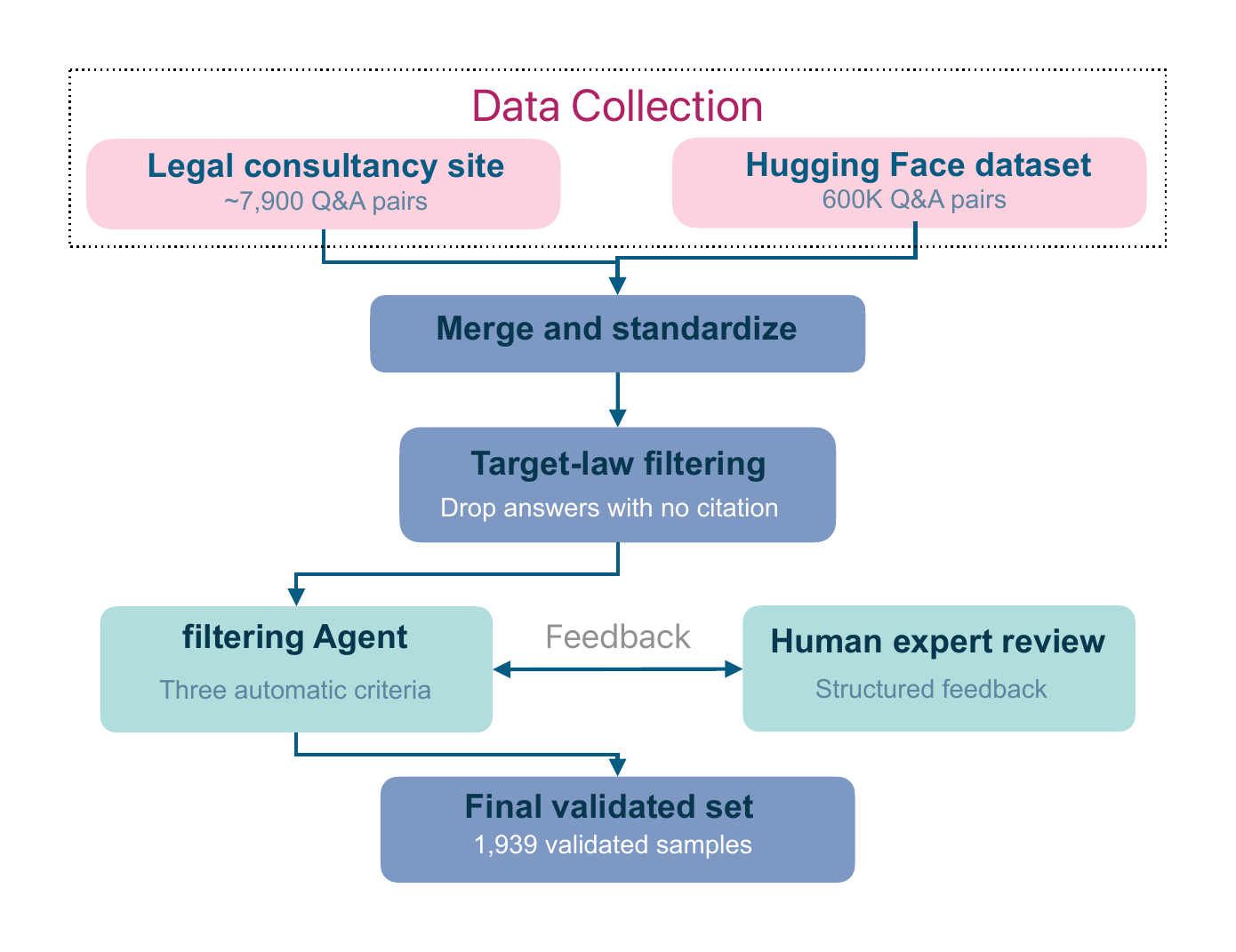}
    \caption{Overview of the TUSI-QA dataset construction pipeline.}
    \label{fig:data_collection_pipeline}
\end{figure}

The first component of TUSI-Bench focuses on real-world legal queries and expert open-ended responses.
The construction of TUSI-QA follows a multi-stage pipeline designed to transform raw legal question-answer interactions into a citation-grounded and quality-controlled benchmark. As illustrated in Figure~\ref{fig:data_collection_pipeline}, the process begins with collecting real-world legal queries and expert responses from multiple sources, followed by data integration and standardization. Subsequently, citation-based filtering is applied to retain only legally grounded responses, and an iterative validation process combining intelligent agent-based evaluation and expert legal review is performed to ensure citation correctness, semantic alignment, and overall legal reliability. The final outcome is a high-quality collection of validated legal question-answer pairs suitable for evaluating retrieval and reasoning capabilities in Legal RAG systems.

\subsubsection{Data Collection and Structural Integration}

In the initial stage, question-answer data was gathered from two public sources to ensure that the samples reflect real-world legal issues. The first source was an Iranian legal consultancy website, where users post questions and receive responses from lawyers. Approximately 7,900 questions along with their corresponding answers were extracted from this source. The data collection was strictly limited to publicly available content, adhering to technical constraints.

The second source was a publicly available dataset published on the Hugging Face platform \cite{iran_legal_persian_qa}, containing 600,000 real Persian legal question-answer pairs with multiple answers from lawyers. This source was used to increase the thematic and stylistic diversity of TUSI-QA.

The data from both sources was standardized and integrated into a uniform structure, ensuring each sample included the question text and the lawyer's response. The output of this stage was a preliminary legal question-answer repository.

\subsubsection{Selection of Target Laws and Explicit References}

After compiling the preliminary dataset, a set of Target Laws was defined to focus the selection of samples on legally-grounded responses. This set of laws includes the primary laws from the bar exam and all relevant legal provisions. These laws serve as both a reference framework for locating legal citations and as a retrieval corpus for subsequent evaluations.

To ensure the quality of the answers, only those responses were selected that explicitly referenced at least one of the target laws, while those without any legal citations were excluded. Once the lawyer responses containing legal references were selected, a step was added to link these responses to the corresponding legal articles. This process involved using an intelligent agent to automatically extract relevant legal provisions from the dataset of laws and link articles to the corresponding answers. Representative examples of the resulting citation-grounded QA instances are shown in Figure~\ref{fig:citation_grounded_qa_examples}.

\subsubsection{Iterative Citation Filtering via Intelligent Agent and Human Evaluation}

Initial qualitative analysis revealed that the mere presence of a legal citation does not guarantee the quality or correctness of the reference. In several cases, cited provisions were partially irrelevant, insufficient for supporting the reasoning, or only loosely connected to the generated answer. In other cases, although legal articles were cited, the reasoning itself was not explicitly grounded in the referenced legal text.

As illustrated in Figure~\ref{fig:data_collection_pipeline}, citation validation for TUSI-QA was therefore formulated as an iterative human-in-the-loop process combining an intelligent agent with expert legal review. In the first stage, the intelligent agent automatically evaluated every sample by jointly analyzing the question, answer, and cited legal provisions according to three criteria: (1) the relevance and sufficiency of the cited legal provisions with respect to the question, (2) explicit grounding of the answer in the cited legal text, and (3) semantic consistency between the question and the generated answer. Samples violating any of these criteria were discarded.

In the second stage, a legal expert inspected a randomly selected subset of the automatically filtered samples. Rather than reviewing the entire dataset, the expert assessed the correctness of the citations, the adequacy of legal grounding, and the consistency between questions and answers. Whenever deficiencies were identified, structured feedback was provided to refine the filtering strategy. The intelligent agent was subsequently updated based on this feedback, the filtering process was repeated, and another random subset was reviewed. This iterative procedure continued until the expert confirmed that the retained samples consistently satisfied both the predefined automatic criteria and the expected legal quality standards.

After the iterative filtering process, the final TUSI-QA dataset contained 1,999 validated question-answer pairs. These samples were subsequently partitioned into training and test sets using a stratified sampling strategy designed to preserve the citation-frequency distribution of legal provisions across both subsets. Instead of performing a purely random split, the sampling process maintained approximately the same proportion of highly frequent and infrequently cited legal articles in the training and test sets. This strategy prevents distributional bias and yields a more representative evaluation benchmark for both retrieval and downstream legal reasoning.

\begin{figure}
    \centering
    \includegraphics[width=1\linewidth]{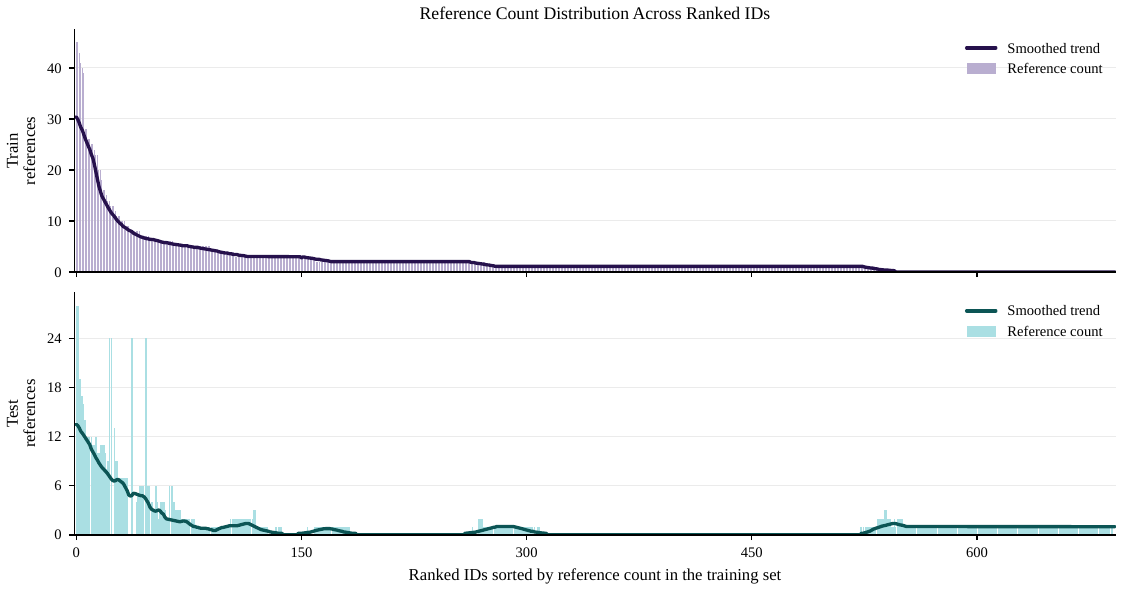}
    \caption{Distribution of citation frequencies across ranked legal provisions in the training and test sets of TUSI-QA. Legal provisions are ranked according to their citation frequency in the training set. The figure shows that the long-tail citation distribution is preserved after the stratified train-test split.}
    \label{fig:reference_distribution}
\end{figure}

Figure~\ref{fig:reference_distribution} illustrates the resulting citation-frequency distributions. The legal references exhibit a pronounced long-tail distribution, where a small number of legal provisions account for a large fraction of all citations, whereas most provisions appear only a few times. Despite this highly skewed distribution, the train-test split preserves a similar citation profile, ensuring that both subsets contain representative proportions of frequent and rare legal provisions.

\begin{figure}
    \centering
    \includegraphics[width=0.6\linewidth]{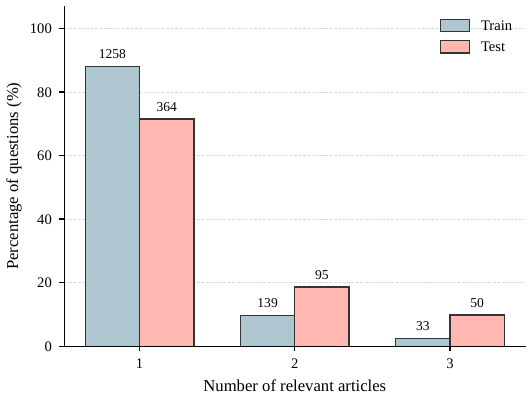}
    \caption{Distribution of the number of relevant legal articles required to answer each question in TUSI-QA (training and test sets). Each question may depend on one, two, or three legal provisions, reflecting different levels of retrieval and reasoning complexity.}
    \label{fig:article_count_distribution}
\end{figure}

Finally, we analyzed the number of supporting legal provisions required to answer each question. Unlike many existing legal question answering datasets that assume a single supporting statute, each question in TUSI-QA may require one, two, or three relevant legal articles. As shown in Figure~\ref{fig:article_count_distribution}, although most questions are grounded in a single legal provision, a considerable number require reasoning over two provisions, and a smaller subset depends on three legal articles. This multi-reference characteristic substantially increases retrieval complexity, as successful systems must retrieve the complete set of supporting legal provisions rather than only the most relevant article, providing a more realistic evaluation of retrieval completeness and legal reasoning capabilities.

\subsubsection{Quantitative Analysis of the Filtering Effect}

To quantitatively evaluate the impact of citation filtering on TUSI-QA, we computed two semantic alignment metrics using BGE-M3 embeddings \cite{chen2024m3}: (1) answer–legal provision similarity and (2) question–legal provision similarity, both measured via cosine similarity.

\begin{figure}
\centering
\includegraphics[width=0.95\textwidth]{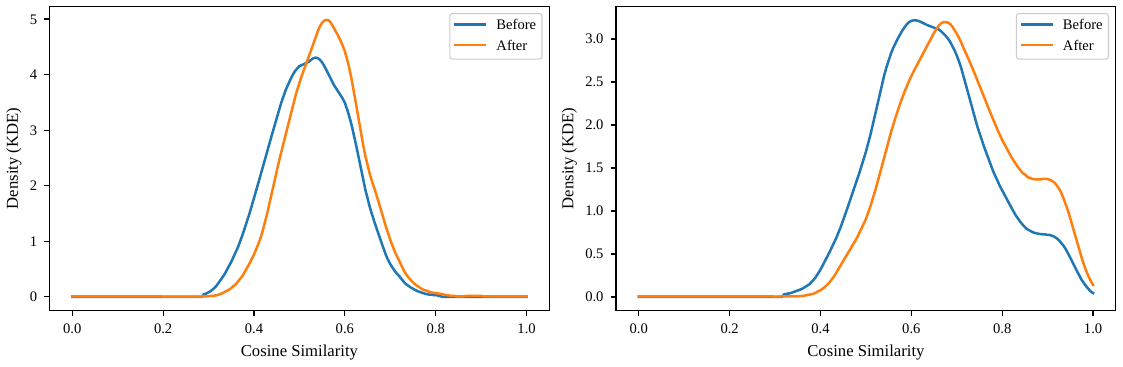}
\caption{KDE plots of cosine similarity scores before and after citation filtering in TUSI-QA. Left: question–legal provision similarity. Right: answer–legal provision similarity. The rightward shift after filtering indicates stronger semantic alignment.}
\label{fig:similarity_before_after}
\end{figure}

As shown in \autoref{fig:similarity_before_after}, the KDE distributions shift to the right after filtering, indicating higher overall semantic alignment. For answer–legal provision similarity, the mean increased from 0.526 to 0.556. A similar improvement was observed for question–legal provision similarity.
To assess whether these differences are statistically meaningful, we conducted Welch's t-test \cite{welch1947}, which is robust to unequal variances and large sample sizes. The test yielded a statistically significant difference for answer–legal provision similarity ($p \ll 0.001$), with a small-to-medium effect size (Cohen's $d = 0.36$). The very small $p$-value indicates that the observed improvement is highly unlikely to have occurred by random variation alone. Comparable statistically significant results were obtained for question–legal provision similarity.

Importantly, while a clear rightward shift is observed, the overall shape of the distributions remains largely unchanged. This suggests that the filtering process did not merely retain samples with extremely high lexical or semantic similarity while discarding the rest. Instead, it selectively reduced low-alignment cases while preserving the natural variability of the dataset.

Overall, the filtering improves semantic grounding and citation relevance without artificially compressing the similarity distribution, resulting in a more stable and realistic component of TUSI-Bench for evaluating Legal RAG and explainable QA systems.

\subsection{TUSI-Bar: Bar Exam Multiple-Choice QA}

The second component of TUSI-Bench, {TUSI-Bar}, consists of multiple-choice questions and answers extracted from the Iranian Bar Exam held over the past three years. Each question in this set follows a four-choice format, with one correct answer and three distractors. Importantly, every question in TUSI-Bar is associated with a specific legal provision, although the annotation is at the law level rather than the individual article level.

The questions, derived from the actual Bar Exam, cover various areas of Iranian law, ensuring thematic diversity. In contrast to TUSI-QA, where each answer directly references a legal article, TUSI-Bar does not provide article-level citations but rather links each question to the broader legal provision or law it addresses. Examples of TUSI-Bar instances, including their answer choices, ground-truth answers, and associated legal sources, are presented in Figure~\ref{fig:bar_exam_dataset_examples}.

The multiple-choice nature of the questions also allows for more controlled evaluation of system performance, as it provides a set of discrete answer options that can be easily compared against model predictions. This structure is ideal for benchmarking, as it allows for clear, quantifiable measures of accuracy based on the system's ability to select the correct answer from the provided choices.

\subsection{Comparison with Existing Datasets}

\begin{figure}[width=\columnwidth,pos=t]
    \centering
    \includegraphics[
        width=\linewidth,
        height=0.8\textheight,
        keepaspectratio
    ]{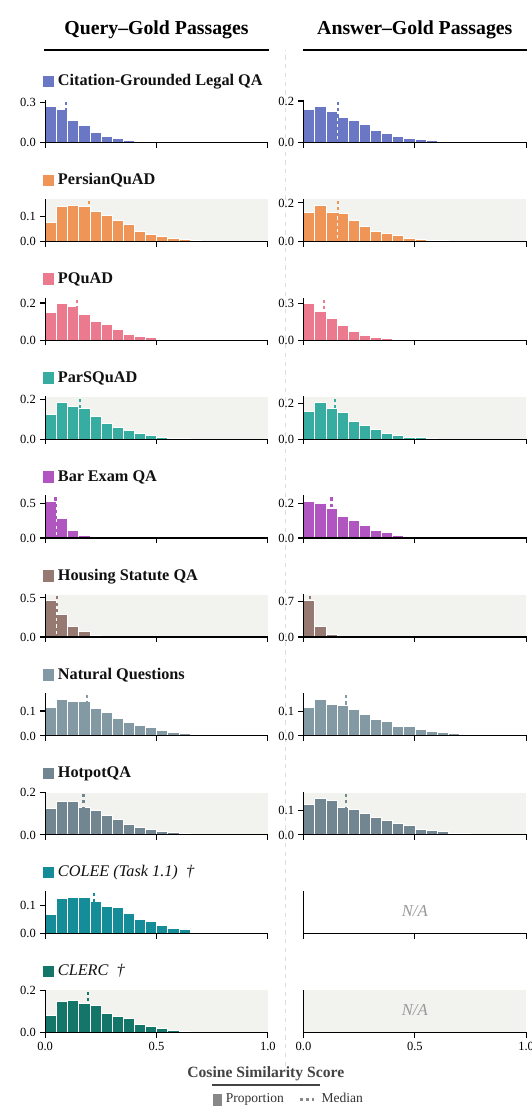}
    \caption{Distribution of cosine similarity scores across datasets for Query--Gold Passage and Answer--Gold Passage settings. The figure illustrates lexical similarity patterns and highlights the distinct behavior of the proposed dataset compared with existing benchmarks.}
    \label{fig:similarity_distributions}
\end{figure}

To better contextualize TUSI-Bench, we compare it with existing legal and general-domain question-answering datasets in terms of scale, passage structure, query and answer length, and lexical overlap. As summarized in Table~\ref{tab:full_comparison}, TUSI-Bench differs from prior resources along multiple dimensions. In terms of scale, it contains fewer passages than several existing datasets, but its passages are relatively longer. In terms of task formulation, unlike datasets that focus primarily on retrieval, TUSI-QA defines a complete open-ended question-answering task. With respect to query and answer structure, it includes long, context-rich questions and explanatory, analytical answers. Finally, as shown in Table~\ref{tab:lexical_similarity} and Figure~\ref{fig:similarity_distributions}, it exhibits a distinct lexical similarity pattern, with a larger gap between users' natural-language descriptions and formal legal texts, while the answers remain grounded in legal sources.

\begin{table}
\centering
\caption{Summary of datasets. We report number of queries (Q), number of passages (P), number of answers (A), and the average length of queries, passages and answers (calculated with the GPT-2 tokenizer \cite{radford2019language}).}
\label{tab:full_comparison}
\begin{tabular}{llrrrrrr}
\toprule
\multirow{2}{*}{Dataset} & \multirow{2}{*}{Split}
& \multicolumn{3}{c}{Total Number}
& \multicolumn{3}{c}{Average Length} \\
\cmidrule(lr){3-5} \cmidrule(lr){6-8}
& & Q & P & A & Q & P & A \\
\midrule

\multirow{2}{*}{TUSI-QA (Ours)}
& Train & 1490 & 7261 & 1490 & 448.04 & 555.22 & 397.50 \\
& Test  & 509  & 7261 & 509  & 426.85 & 548.42 & 362.95 \\
\midrule

\multicolumn{8}{l}{\textbf{Comparison}} \\

\multirow{2}{*}{PersianQuAD \cite{persianquad2022}}
& Train & 18567 & 3149 & 18567 & 59.07  & 1032.47 & 16.14 \\
& Test  & 1000  & 35   & 2339  & 57.79  & 1525.66 & 12.05 \\
\midrule

\multirow{2}{*}{PQuAD \cite{darvishi2023pquad}}
& Train & 63994 & 8979 & 48273 & 65.43 & 782.99 & 33.63 \\
& Test  & 8002  & 1059 & 8540  & 67.93 & 771.51 & 33.13 \\
\midrule

\multirow{2}{*}{ParsQuAD \cite{Abadani2021ParSQuAD}}
& Train & 18906 & 5723 & 15517 & 59.36 & 582.09 & 11.20 \\
& Test  & 5711  & 1165 & 8742  & 59.53 & 762.67 & 12.59 \\
\midrule

Bar Exam QA \cite{zheng2025legal}
& -- & 1,195 / 1,815 & 856,835 & -- & 172 / 157 & 131 & -- \\

Housing Statute QA \cite{zheng2025legal}
& -- & 6,853 & 1,837,403 & -- & 15 & 349 & -- \\

Natural Questions \cite{kwiatkowski2019natural}
& -- & 3,452 & 2,681,468 & -- & 11 & 102 & -- \\

HotpotQA \cite{yang2018hotpotqa}
& -- & 7,405 & 5,233,329 & -- & 20 & 63 & -- \\

COLIEE \cite{goebel2024coliee}
& -- & 1,278 & 5,616 & -- & 6,730 & 6,768 & -- \\

CLERC \cite{hou2025clerc}
& -- & 2,851 & 1,842,422 & -- & 415 & 3,303 & -- \\

\bottomrule
\end{tabular}
\end{table}

\begin{table}
\centering
\caption{Lexical similarity scores for the retrieval task and downstream QA task across datasets. Values are reported as mean $\pm$ 95\% confidence interval.}
\label{tab:lexical_similarity}
\begin{tabular}{lcc}
\toprule
\textbf{Dataset} & \textbf{LS\textsubscript{retrieval}} & \textbf{LS\textsubscript{QA}} \\
\midrule
TUSI-QA (Ours) & $0.03 \pm 0.00$ & $0.19 \pm 0.02$ \\
PersianQuAD \cite{persianquad2022} & $0.12 \pm 0.00$ & $0.09 \pm 0.00$ \\
PQuAD \cite{darvishi2023pquad} & $0.16 \pm 0.00$ & $0.12 \pm 0.00$ \\
ParSQuAD \cite{Abadani2021ParSQuAD} & $0.10 \pm 0.00$ & $0.12 \pm 0.00$ \\
\midrule
Bar Exam QA \cite{zheng2025legal} & $0.07 \pm 0.00$ & $0.14 \pm 0.01$ \\
Housing Statute QA \cite{zheng2025legal} & $0.09 \pm 0.00$ & $0.00 \pm 0.00$ \\
Natural Questions \cite{kwiatkowski2019natural} & $0.27 \pm 0.01$ & $0.25 \pm 0.01$ \\
HotPotQA \cite{yang2018hotpotqa} & $0.26 \pm 0.00$ & $0.24 \pm 0.00$ \\
COLIEE \cite{goebel2024coliee} & $0.27 \pm 0.00$ & -- \\
CLERC \cite{hou2025clerc} & $0.26 \pm 0.01$ & -- \\
\bottomrule
\end{tabular}
\end{table}

In terms of scale, the number of passages in TUSI-Bench is smaller than in some legal benchmarks, such as Bar Exam QA and Housing Statute QA \cite{zheng2025legal}. However, as shown in Table~\ref{tab:full_comparison}, the average passage length in TUSI-QA is higher than in these two benchmarks. Thus, although the retrieval corpus is smaller, its textual units are longer and more content-rich. By contrast, datasets such as COLIEE \cite{goebel2024coliee} and CLERC \cite{hou2025clerc} have a different task formulation and contain both a substantial number of passages and considerably longer average passage lengths. The long query lengths in these datasets are mainly due to the structure of legal retrieval tasks, where long legal texts or case descriptions are used as retrieval inputs, rather than natural user questions in a question-answering framework.

From the perspective of language and QA formulation, PersianQuAD \cite{persianquad2022}, PQuAD \cite{darvishi2023pquad}, and ParsQuAD \cite{Abadani2021ParSQuAD} are among the main Persian resources for evaluating question-answering and RAG systems in the general domain. These datasets mainly consist of questions whose answers are short, specific, and usually extractable from the reference text. This characteristic is also reflected in the answer-length statistics in Table~\ref{tab:full_comparison}: the average answer length in these datasets is substantially lower than in TUSI-QA. Therefore, although these resources are valuable for evaluating Persian information retrieval and answer extraction, they generally do not require the generation of long, analytical, or explanatory answers.

In contrast, TUSI-QA is designed as an open-ended QA benchmark. Its answers are not merely short or factoid responses; rather, they require legal explanation, interpretation of statutory provisions, and alignment between the cited law and the conditions described in the question. This distinction is reflected in Table~\ref{tab:full_comparison}, where the answers in both the training and test splits are considerably longer than those in Persian QuAD-based datasets. Moreover, among the QA datasets considered in this study, TUSI-QA stands out in terms of both query and answer length. This reflects its focus on context-dependent legal problems, where users often need to describe the facts of the case, temporal conditions, relationships between parties, and relevant legal details. Although COLIEE has the highest average query length statistically, its task formulation differs from QA datasets: its queries are not necessarily natural-language questions and often correspond to legal cases or text segments. Therefore, its query length is not directly comparable to the question length in QA datasets.

TUSI-QA also exhibits distinctive lexical similarity characteristics. As shown in Table~\ref{tab:lexical_similarity} and Figure~\ref{fig:similarity_distributions}, its lexical similarity in the retrieval setting is lower than that of the other compared datasets. This indicates limited surface-level overlap between user questions and the corresponding legal passages, which is consistent with the nature of legal consulting scenarios: users typically describe their problems in natural and narrative language, whereas statutory texts are written in a formal and structured style. In contrast, the higher lexical similarity in the QA setting indicates that the final answers remain connected to the relevant legal texts and do not drift away from legal grounding.

Overall, TUSI-Bench is not necessarily the largest resource in terms of the number of passages, but it offers a distinctive combination of properties: relatively long passages, context-rich user questions, long open-ended answers, and explicit grounding in legal texts. In addition, among the QA datasets considered in this study, TUSI-QA is particularly prominent in terms of average query and answer length. These characteristics make TUSI-Bench a suitable benchmark for evaluating Legal RAG systems in scenarios where both accurate evidence retrieval and explainable answer generation are essential.

For all experiments, retrieval is performed over the complete collection of legal provisions introduced in Section~\ref{sec:knowledge_layer}, and all evaluations are conducted on the corresponding test splits of TUSI-Bench.

\section{Experimental Setup and Results}
\label{sec:results}

This section presents the experimental results of the proposed benchmark. We consider three evaluation tasks: retrieval of relevant legal provisions, open-ended question answering, and legal reasoning. Retrieval and open-ended QA are evaluated on the proposed Legal QA benchmark, while legal reasoning is evaluated on the Bar Exam benchmark as a four-choice question answering task.


All experiments were conducted on a workstation equipped with an NVIDIA RTX A6000 GPU. For large language model inference, including the iterative citation filtering and quality control pipeline described in Section~\ref{sec:benchmark}, we employ Gemma 3 27B \cite{team2025kimi} quantized to 8-bit precision. Document embeddings are indexed using the Weaviate vector database, and retrieval is performed using dense, BM25, or hybrid search depending on the experimental setting. Unless otherwise specified, all reported results are obtained using identical preprocessing and retrieval settings for fair comparison.

\subsection{Evaluation Metrics}

The proposed benchmark supports the evaluation of both Legal Information Retrieval (LIR) and Legal Question Answering (LQA). Accordingly, we employ two complementary evaluation protocols. For retrieval experiments, we adopt standard information retrieval metrics that assess the quality of document ranking. For question answering, we employ an LLM-as-a-Judge framework to evaluate the legal quality and practical usefulness of generated responses.

\paragraph{Retrieval Metrics}

Let $Q$ denote the set of evaluation queries and let $R_q$ be the set of relevant documents associated with query $q$.

\paragraph{Precision@k.}
Precision measures the proportion of retrieved documents that are relevant among the top-$k$ retrieved results:
\begin{equation}
P@k=\frac{|Rel_k|}{k},
\end{equation}
where $Rel_k$ denotes the set of relevant documents retrieved within the first $k$ positions. Precision reflects the ability of a retrieval system to avoid returning irrelevant legal provisions.

\paragraph{Recall@k.}
Recall measures the proportion of relevant documents that are successfully retrieved:
\begin{equation}
R@k=\frac{|Rel_k|}{|R_q|}.
\end{equation}

Recall is particularly important in legal retrieval since missing a relevant statutory provision may result in incomplete legal reasoning.

\paragraph{Mean Average Precision (MAP).}
For a query $q$, Average Precision (AP) is defined as
\begin{equation}
AP(q)=
\frac{1}{|R_q|}
\sum_{k=1}^{N}
P@k \times rel(k),
\end{equation}
where $N$ is the total number of retrieved documents and $rel(k)$ equals 1 if the document at rank $k$ is relevant and 0 otherwise.
The Mean Average Precision is computed as
\begin{equation}
MAP=
\frac{1}{|Q|}
\sum_{q\in Q}
AP(q).
\end{equation}
MAP rewards systems that consistently rank relevant legal documents near the top of the retrieval list.

\paragraph{Mean Reciprocal Rank (MRR).}
MRR \cite{voorhees1999trec8} evaluates the rank of the first relevant retrieved document:
\begin{equation}
MRR=
\frac{1}{|Q|}
\sum_{q\in Q}
\frac{1}{rank_q},
\end{equation}
where $rank_q$ denotes the position of the first relevant document for query $q$. MRR is particularly useful in legal retrieval scenarios where identifying a key statutory provision early can significantly improve downstream reasoning.

\paragraph{Normalized Discounted Cumulative Gain (NDCG).}
NDCG \cite{jarvelin2002cumulated} measures ranking quality while assigning greater importance to highly ranked relevant documents.
The Discounted Cumulative Gain is defined as
\begin{equation}
DCG@k=
\sum_{i=1}^{k}
\frac{2^{rel_i}-1}
{\log_2(i+1)},
\end{equation}
where $rel_i$ denotes the relevance score of the document at position $i$.
The normalized version is
\begin{equation}
NDCG@k=
\frac{DCG@k}{IDCG@k},
\end{equation}
where $IDCG@k$ is the maximum achievable DCG. NDCG is well suited for legal retrieval because multiple statutory provisions may be relevant to a single legal question, and their ranking order is important.

\paragraph{LLM-as-a-Judge Evaluation}

While retrieval metrics evaluate the ability to identify relevant legal documents, they cannot fully assess the quality of generated legal responses. Therefore, for the Legal Question Answering task, we adopt an LLM-as-a-Judge \cite{zheng2023judging} evaluation framework.

We employ DeepSeek-V4-Pro \cite{deepseek2026v4series} as an automatic evaluator. Given the user question, the retrieved legal context, the reference lawyer answer, and the model-generated answer, the judge model produces both a qualitative justification and a score for each evaluation criterion.

The generated answers are evaluated according to five complementary dimensions, each scored on a five-point integer scale.

\begin{itemize}
    \item \textbf{Correctness}: legal correctness and consistency with the reference answer and supporting legal evidence.
    \item \textbf{Effectiveness}: practical usefulness and the ability to guide the user toward an appropriate legal course of action.
    \item \textbf{Objectiveness}: professional and impartial legal tone while avoiding unsupported certainty.
    \item \textbf{Clarity}: readability, organization, and accessibility of legal concepts.
    \item \textbf{Completeness}: adequate coverage of legally relevant aspects without introducing unnecessary or unrelated content.
\end{itemize}
Each criterion receives an integer score from 1 to 5, where higher scores indicate better performance. In addition to the individual criterion scores, the overall QA performance is reported as the average score across all five dimensions. This evaluation protocol enables assessment not only of factual legal correctness but also of the practical, professional, and communicative quality of generated legal advice.

\begin{figure}
\centering
\includegraphics[width=0.95\textwidth]{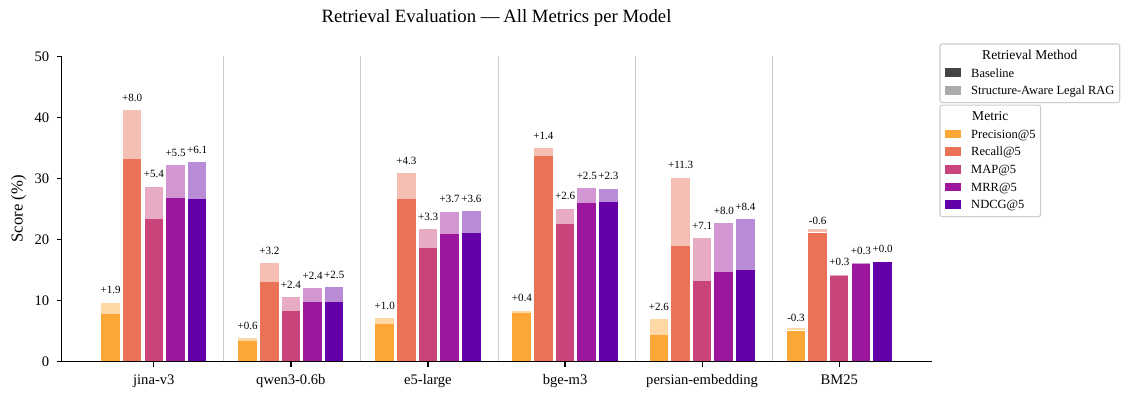}
\caption{Retrieval evaluation results across different embedding models and the proposed Structure-Aware Legal RAG framework. The figure illustrates consistent improvements over the baseline in most configurations.}
\label{fig:retrieval_results}
\end{figure}

\subsection{Baselines}

We evaluate two primary retrieval baselines: BM25 \cite{robertson2009bm25} and a vanilla dense retrieval-based RAG pipeline~\cite{lewis2020rag}. BM25 serves as a sparse lexical matching baseline, while the dense retrieval setting encodes each segmented legal chunk into vector representations stored in a vector database, where relevant passages are retrieved based on semantic similarity with the input query.

To ensure a fair comparison across embedding strategies, we evaluate the dense retrieval pipeline using five different embedding models: \texttt{jina-embeddings-v3}\allowbreak{}~\cite{sturua2024jina}, \texttt{bge-m3}~\cite{chen2024m3}, \texttt{multilingual-e5-large}\allowbreak{}~\cite{wang2024multilinguale5}, and a domain-adapted Persian embedding model based on fine-tuning of a pre-trained multilingual XLM-RoBERTa encoder, trained on Persian legal corpora, and \texttt{Qwen3-Embedding-0.6B}\allowbreak{}~\cite{zhang2025qwen3embedding}. These models are selected to cover both multilingual and Persian-specialized embedding spaces, enabling a comprehensive evaluation of retrieval performance across different representation paradigms.

In particular, the first three models are widely adopted in prior Persian and multilingual retrieval studies due to their strong semantic retrieval performance~\cite{zinvandi2025famteb}, while the Persian-specific model is included as a domain-adapted baseline for Persian legal text retrieval. Finally, \texttt{Qwen3-Embedding-0.6B}\allowbreak{} is added as a recently released large-scale multilingual embedding model from the Qwen3 family, representing a stronger modern baseline.

\subsection{Structure-Aware Legal RAG}

Recent studies on Graph Retrieval-Augmented Generation show that conventional RAG is not always sufficient for specialized domains~\cite{edge2024graphrag}. In such domains, relevant knowledge is often distributed across multiple documents, and understanding it depends on conceptual and structural relations among knowledge units. In this setting, flat text retrieval may fail to preserve the structural dependencies required for domain-specific reasoning. This limitation is particularly important in legal question answering, where the correct answer may depend not only on the semantic similarity between a query and a statutory provision, but also on the relation of that provision to other articles, authoritative interpretations, and Judicial Unification Rulings.

To address this issue, we propose a structure-aware legal retrieval method that combines semantic retrieval with graph-based legal context expansion. The method uses the legal knowledge graph introduced in our dataset. In this graph, nodes represent statutory articles, expert interpretations, and Judicial Unification Rulings, while edges represent legal relations such as article-to-article references, article--interpretation links, and article--ruling associations.

In the proposed method, all legal texts are first segmented into semantically coherent chunks. Each chunk is then encoded as a dense vector and stored in a vector database. Given a user query, the system first retrieves the chunks that are most semantically similar to the query. This step provides the initial evidence required for answering the question.

Next, the retrieved chunks are mapped to their corresponding nodes in the legal knowledge graph. For each matched node, the method retrieves all nodes located in its one-hop neighborhood. The resulting one-hop expanded node set is returned as the graph-augmented retrieval output and is used as the evidence set for downstream answer generation.

As a result, the final evidence set includes not only the chunks that are semantically similar to the query, but also the legally connected articles and sources. This structural expansion allows the system to retrieve materials that may be legally necessary for reasoning, even if they are not ranked highly by semantic similarity alone.

Finally, the expanded evidence set is provided to the language model as grounded context, and the model generates the answer based on the retrieved statutory articles, interpretations, and judicial decisions. In this way, the proposed method combines semantic similarity with the explicit structure of legal knowledge, making retrieval more aligned with statute-based legal question answering than flat retrieval.

\begin{table}[width=0.8\textwidth,pos=t]
\centering
\caption{Downstream evaluation results on the Cited Question-Answer Set.}
\label{tab:cited_qa_results}
\begin{tabular}{lccccc}
\toprule
\textbf{Method} & \textbf{Correctness} & \textbf{Effectiveness} & \textbf{Objectiveness} & \textbf{Clarity} & \textbf{Completeness} \\
\midrule
Gemma3 27B 8bit~\cite{team2025kimi}
& 1.947 & 2.092 & 3.314 & \underline{3.827} & 1.782 \\

Gemma3 27B 8bit~\cite{team2025kimi} + Structure-Aware \\Legal RAG
& \textbf{2.420} & \textbf{2.540} & \underline{3.487} & 3.644 & \textbf{2.206} \\

GPT-5-nano~\cite{singh2025openai}
& \underline{2.110} & \underline{2.450} & \textbf{4.150} & \textbf{4.000} & \underline{2.010} \\
\bottomrule
\end{tabular}
\end{table}

\subsection{Retrieval Evaluation}

In the first stage, we evaluate the retrieval performance of the proposed \textit{Structure-Aware Legal RAG} framework in comparison with the BM25 \cite{robertson2009bm25} baseline and embedding-based retrieval models, as shown in Figure~\ref{fig:retrieval_results}. The objective is to assess the impact of incorporating structural legal information and relational signals into the retrieval process.

The results indicate that the proposed framework improves retrieval performance across all embedding-based configurations. The only exception is BM25, where a decrease is observed in both Precision@5 and Recall@5, dropping by 0.6 and 0.3 percentage points, respectively, as reported in Figure~\ref{fig:retrieval_results}. This suggests that purely lexical matching methods are limited in capturing the semantic and structural complexity of legal documents.

Among embedding-based models, Jina-v3 achieves the best performance, reaching a Recall@5 of 41.22\% with the proposed Structure-Aware Legal RAG framework, an improvement of up to 8 percentage points over its dense-retrieval-only counterpart, as shown in Figure~\ref{fig:retrieval_results}. This demonstrates the effectiveness of combining dense embeddings with structure-aware retrieval for improving relevant passage identification.

Overall, these results highlight the importance of structural signals in legal retrieval, particularly in scenarios where lexical similarity between queries and legal provisions is low. As illustrated in Figure~\ref{fig:retrieval_results}, relational and structural cues play a critical role in bridging the semantic gap between queries and legal texts.

\subsection{Downstream Evaluation}

To comprehensively evaluate the proposed framework, we conduct experiments on two downstream legal reasoning tasks: (i) a Cited Question Answering task, which requires generating open-ended, citation-grounded legal responses, and (ii) a Bar Exam multiple-choice question answering task, designed to assess structured legal reasoning under constrained answer selection.

In the Cited QA setting, model performance is evaluated across five dimensions, including correctness, effectiveness, objectiveness, clarity, and completeness. As shown in Table~\ref{tab:cited_qa_results}, the Structure-Aware Legal RAG framework improves performance on four of the five evaluation criteria compared to the non-RAG baseline, with the largest gains observed in correctness, effectiveness, and completeness. In particular, incorporating structured legal knowledge leads to more reliable and better-grounded responses, improving both reasoning quality and factual consistency. Clarity is the one criterion that decreases with the addition of retrieved context (from 3.827 to 3.644), a plausible trade-off in which longer, evidence-grounded answers become somewhat less concise and readable even as they become more accurate and complete. Objectiveness shows a modest improvement, indicating that grounding responses in retrieved legal sources does not come at the cost of a professional and impartial tone.
For additional reference, GPT-5-nano~\cite{singh2025openai} is also evaluated in a zero-shot setting without access to the proposed retrieval framework. Notably, the Structure-Aware Legal RAG framework still yields improvements over the open-source Gemma3 baseline, highlighting the benefits of incorporating structured legal knowledge even with an open-source model.

In the Bar Exam multiple-choice setting, we further evaluate the model's ability to perform structured legal reasoning and select the correct option among four candidates. As reported in Table~\ref{tab:downstream_results}, the proposed Structure-Aware Legal RAG framework consistently outperforms both the vanilla model and the standard RAG baseline, improving accuracy from 33.0\% to 37.7\%. These results demonstrate that structure-aware retrieval improves not only open-ended generation tasks but also strengthens discriminative reasoning in exam-style legal problems, leading to more robust and reliable decision-making.

\begin{table}[width=0.8\textwidth,pos=t]
\centering
\caption{Downstream evaluation results on Bar Exam multiple-choice QA.}
\label{tab:downstream_results}

\begin{tabular*}{0.9\textwidth}{@{\extracolsep{\fill}}lc}
\toprule
\textbf{Method} & \textbf{Accuracy (\%)} \\
\midrule
Gemma3 27B 8bit~\cite{team2025kimi} & 33.0 \\
Gemma3 27B 8bit~\cite{team2025kimi} + RAG & 36.8 \\
Gemma3 27B 8bit~\cite{team2025kimi} + Structure-Aware Legal RAG & \underline{37.7} \\
Deepseek V4 flash~\cite{deepseek2026v4series} & \textbf{57.5} \\
\bottomrule
\end{tabular*}
\end{table}

\subsection{Discussion}
\label{sec:discussion}

The proposed work addresses a critical gap in legal information retrieval and question answering for low-resource languages, particularly Persian. Existing benchmarks in this domain are either synthetic, limited in scope, or lack explicit grounding in authoritative legal structures. In contrast, this study introduces a multi-layered Persian legal dataset that integrates statutes, legal texts, doctrinal writings, interpretative sources, and judicial decisions into a unified and structured framework.

The experimental results demonstrate that leveraging structured legal knowledge significantly improves both retrieval and reasoning performance. The observed improvements are primarily attributed to the explicit modeling of legal relationships through a knowledge graph, which enables the retrieval system to move beyond surface-level semantic similarity and incorporate legally meaningful dependencies.

Furthermore, the introduction of two complementary components under {TUSI-Bench}---{TUSI-QA}, an open-ended citation-grounded question answering task, and \textbf{TUSI-Bar}, a multiple-choice legal reasoning exam---provides a more realistic and comprehensive evaluation setting for Persian legal NLP systems. These benchmarks directly address the limitations of existing datasets, which typically fail to capture real-world legal reasoning complexity or lack reliable ground truth alignment with authoritative sources.

Despite these advantages, the proposed framework also introduces certain challenges. The construction of a high-quality legal knowledge graph requires careful curation and expert validation, as incorrect or incomplete relational links may propagate noise into the retrieval process. Additionally, evaluating open-ended legal answers remains inherently challenging due to the complexity of legal interpretation and variability in acceptable responses.

\section{Conclusion}
\label{sec:conclusion}

This study introduced TUSI-Bench, a Persian legal benchmark for evaluating retrieval-augmented generation systems, together with the structured legal knowledge layer on which it is built. The knowledge layer comprises 7,261 articles from 50 statutes, linked to expert interpretations and Supreme Court Unification Rulings through a legal knowledge graph of 9,272 nodes and 5,183 cross-references. On top of this layer, TUSI-Bench provides two complementary tasks: TUSI-QA, an open-ended question answering benchmark of 1,999 questions, and TUSI-Bar, a four-option multiple-choice benchmark derived from Iranian Bar examination questions.

The main novelty of TUSI-Bench lies in the nature and quality of its data. TUSI-QA is built from real lawyer--client consultations rather than synthetically generated questions, so its queries reflect the narrative, fact-rich way in which people actually describe legal problems, and they exhibit very low lexical overlap with the governing statutory text. Each retained answer is explicitly linked to one to three supporting articles of the target laws, which makes citation-level evaluation of both retrieval and generation possible. The data were cleaned through an iterative filtering process that combines an intelligent agent with expert legal review, discarding samples whose citations were irrelevant, insufficient, or not actually used in the reasoning; the resulting semantic alignment between questions, answers, and cited provisions is significantly higher after filtering. Finally, the train--test split preserves the long-tail citation distribution of the legal provisions, and all questions are linked to a shared retrieval corpus, so results are comparable across systems.

Our experiments show that incorporating structured legal knowledge helps. A simple structure-aware baseline, which expands retrieved chunks with their one-hop neighbors in the legal knowledge graph, improves Recall@5 by up to 8 percentage points over its dense-retrieval counterpart, raises answer quality on TUSI-QA, and increases accuracy on TUSI-Bar. At the same time, the absolute scores remain low, which indicates that TUSI-Bench is a challenging benchmark with substantial room for improvement.

Several directions remain for future work. First, we plan to expand the corpus beyond the current 50 statutes by adding further laws and other legal documents, such as executive regulations, additional judicial rulings, and legal commentaries, which would broaden coverage and make the benchmark more representative of legal practice. Second, the current graph is unweighted: all edges of a given type contribute equally, and one-hop expansion can therefore introduce noise, especially around highly connected hub articles. Assigning weights to edges, for example based on reference frequency, relation type, semantic similarity, or co-citation patterns in TUSI-QA, would allow the graph to model the strength of the relations between provisions more faithfully and to rank expanded neighbors rather than include them all. Third, we plan to improve automatic relation extraction to enhance the scalability and coverage of the graph, and to extend TUSI-Bar with article-level annotations in addition to the current law-level links.

\noindent\textbf{Data Availability}
The dataset developed in this study is publicly available in the
\href{https://huggingface.co/datasets/Marykka/TUSI-Bench}{TUSI-Bench repository}
on the Hugging Face Hub.


\bibliographystyle{unsrt}

\bibliography{cas-refs}
liography{cas-refs}
\appendix
\section*{Appendix}

\renewcommand{\thefigure}{\Alph{section}.\arabic{figure}}
\setcounter{figure}{0}

\section{Examples from TUSI-Bench}
\label{app:plex_examples}

This appendix provides qualitative examples from the proposed {TUSI-Bench}
evaluation suite. The examples illustrate the structure and content of the two
introduced benchmarks, {TUSI-QA} and {TUSI-Bar}. Specifically,
TUSI-QA is designed as a citation-grounded open-ended legal question answering
benchmark, where each instance connects a legal query with an answer supported
by authoritative legal sources. TUSI-Bar, on the other hand, evaluates legal
reasoning ability through multiple-choice questions derived from professional
legal examination settings.

\begin{figure}[ht]
    \centering

    \begin{subfigure}[b]{0.95\linewidth}
        \centering
        \includegraphics[width=\linewidth]{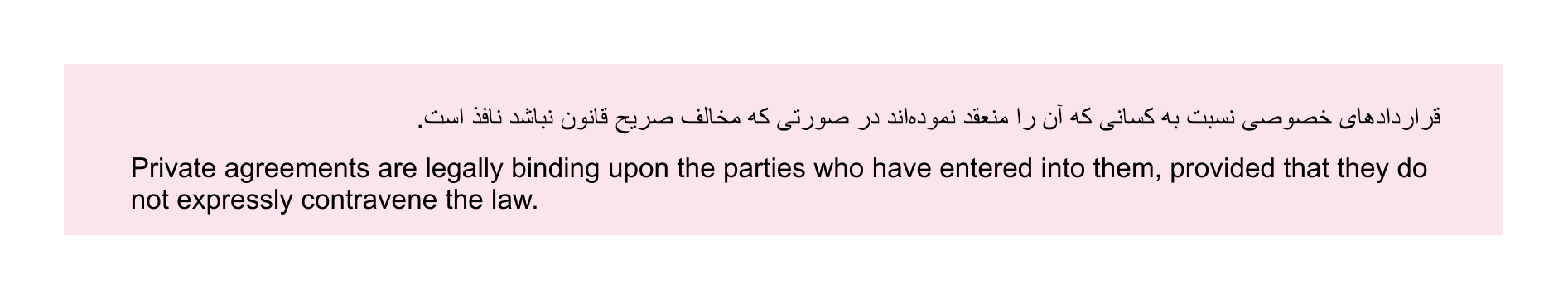}
        \caption{Original legal article (Article 10 of the Iranian Civil Code).}
        \label{fig:article10_a}
    \end{subfigure}
    \hfill
    \begin{subfigure}[b]{0.95\linewidth}
        \centering
        \includegraphics[width=\linewidth]{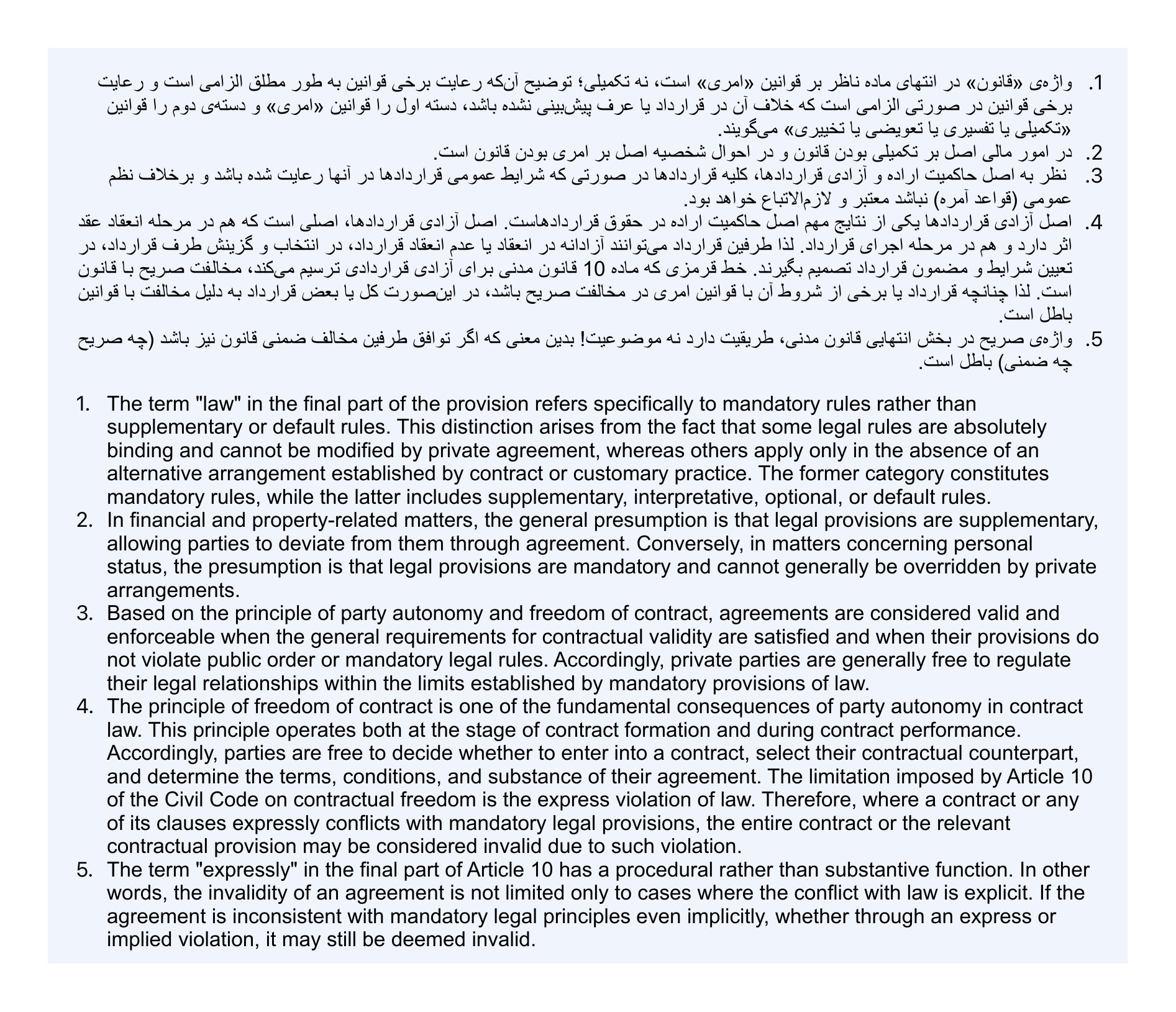}
        \caption{Corresponding expert interpretation of Article 10.}
        \label{fig:article10_b}
    \end{subfigure}

    \caption{
    An example instance from TUSI-QA, consisting of
    a statutory article and its corresponding expert interpretation.
    }
    \label{fig:article10_structure}
\end{figure}

Figure~\ref{fig:article10_structure} presents an example instance from
TUSI-QA, consisting of a statutory provision from the Iranian legal system and
its corresponding expert interpretation. Such paired representations enable
the evaluation of whether models can retrieve and reason over relevant legal
texts beyond surface-level matching.

\begin{figure}[ht]
    \centering

    \begin{subfigure}[b]{0.95\linewidth}
        \centering
        \includegraphics[width=\linewidth]{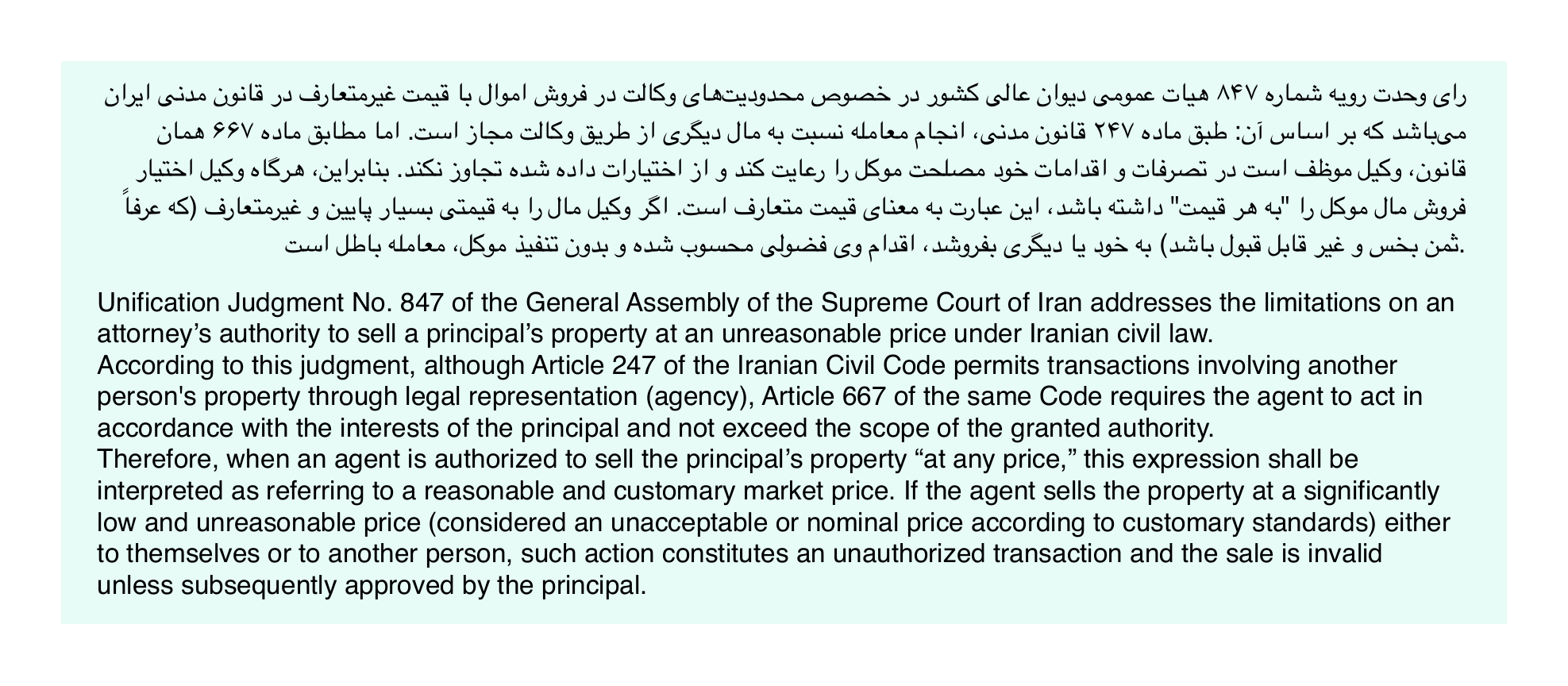}
        \caption{
        Summary of Unification Judgment No. 847 of the General Assembly of the Supreme Court of Iran,
        issued on May 14, 2024.
        }
        \label{fig:verdict847}
    \end{subfigure}

    \vspace{0.5cm}

    \begin{subfigure}[b]{0.95\linewidth}
        \centering
        \includegraphics[width=\linewidth]{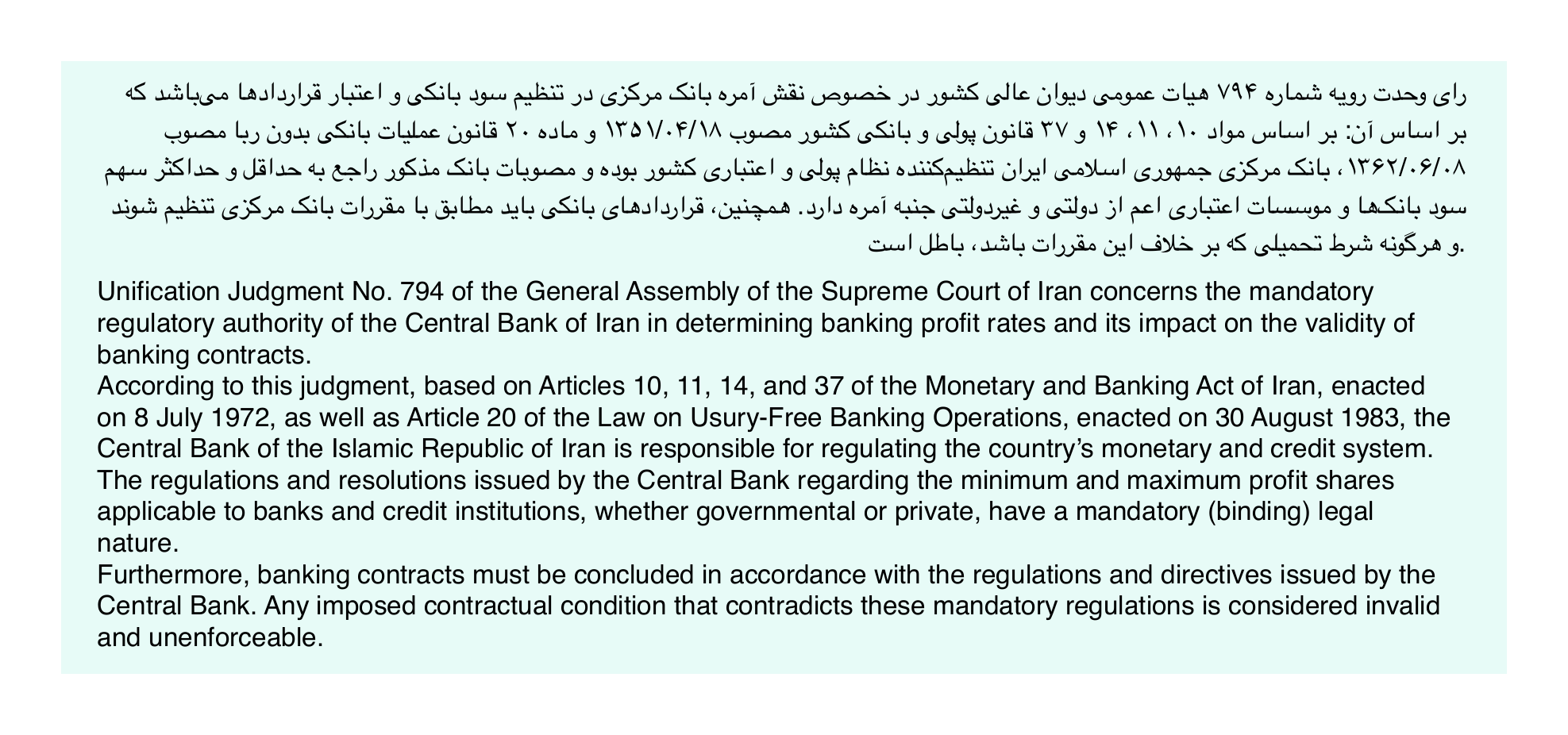}
        \caption{
        Summary of Unification Judgment No. 794 of the General Assembly of the Supreme Court of Iran,
        issued on August 11, 2020.
        }
        \label{fig:verdict794}
    \end{subfigure}

    \caption{
    Examples of judicial precedents included in TUSI-QA:
    Unification Judgments No. 847 and No. 794 of the General Assembly
    of the Supreme Court of Iran. Each example presents a concise summary
    of the corresponding judicial decision.
    }
    \label{fig:unification_judgments_examples}
\end{figure}

Figure~\ref{fig:unification_judgments_examples} shows examples of judicial
precedents included in TUSI-QA. These examples demonstrate the incorporation
of Supreme Court unification judgments, which represent an important source of
legal reasoning and interpretation in the Iranian judicial system.

\begin{figure}[ht]
    \centering

    \begin{subfigure}[b]{0.95\linewidth}
        \centering
        \includegraphics[width=\linewidth]{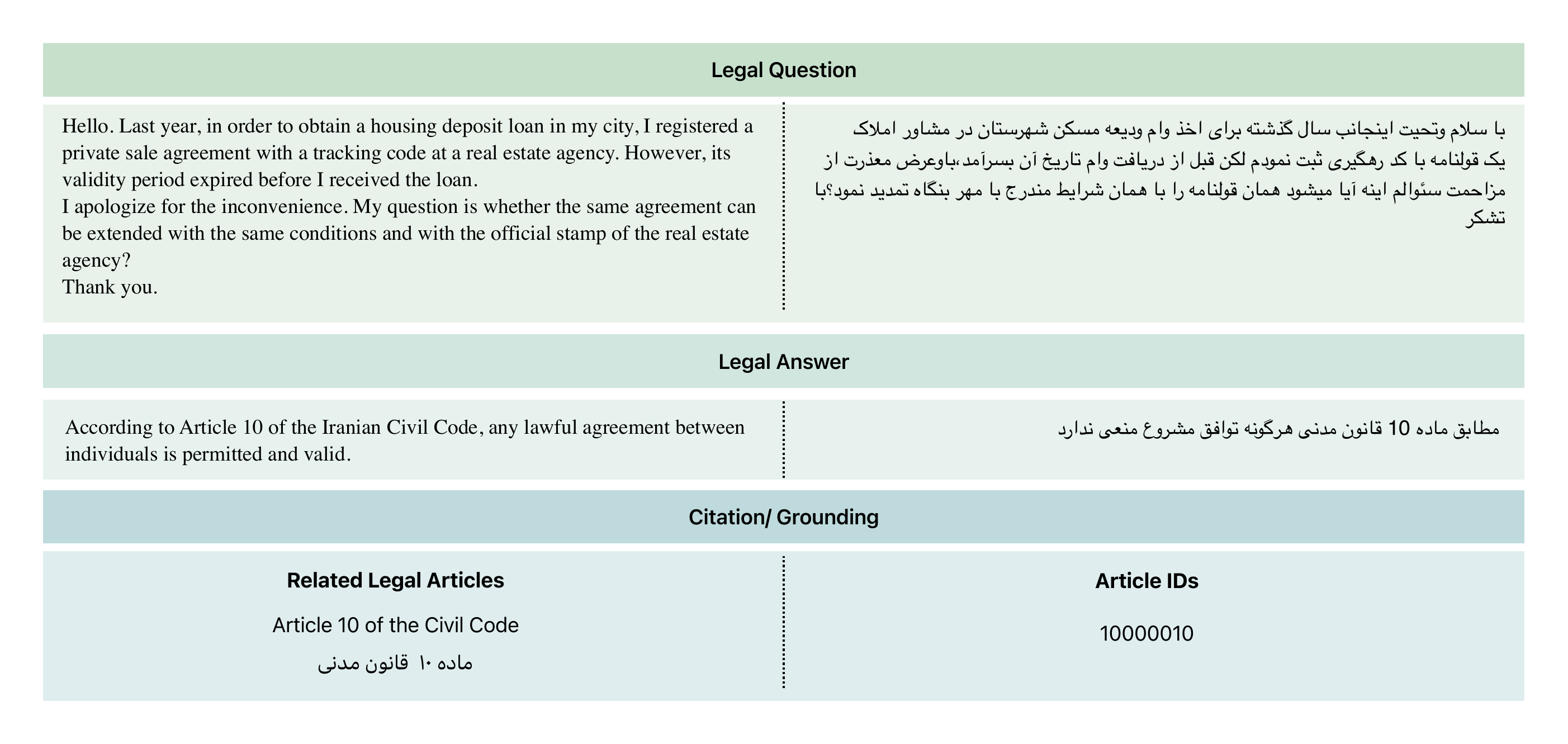}
        \caption{Example instance 1 from TUSI-QA.}
        \label{fig:qa_example_1}
    \end{subfigure}

    \vspace{0.4cm}



    \begin{subfigure}[b]{0.95\linewidth}
        \centering
        \includegraphics[width=\linewidth]{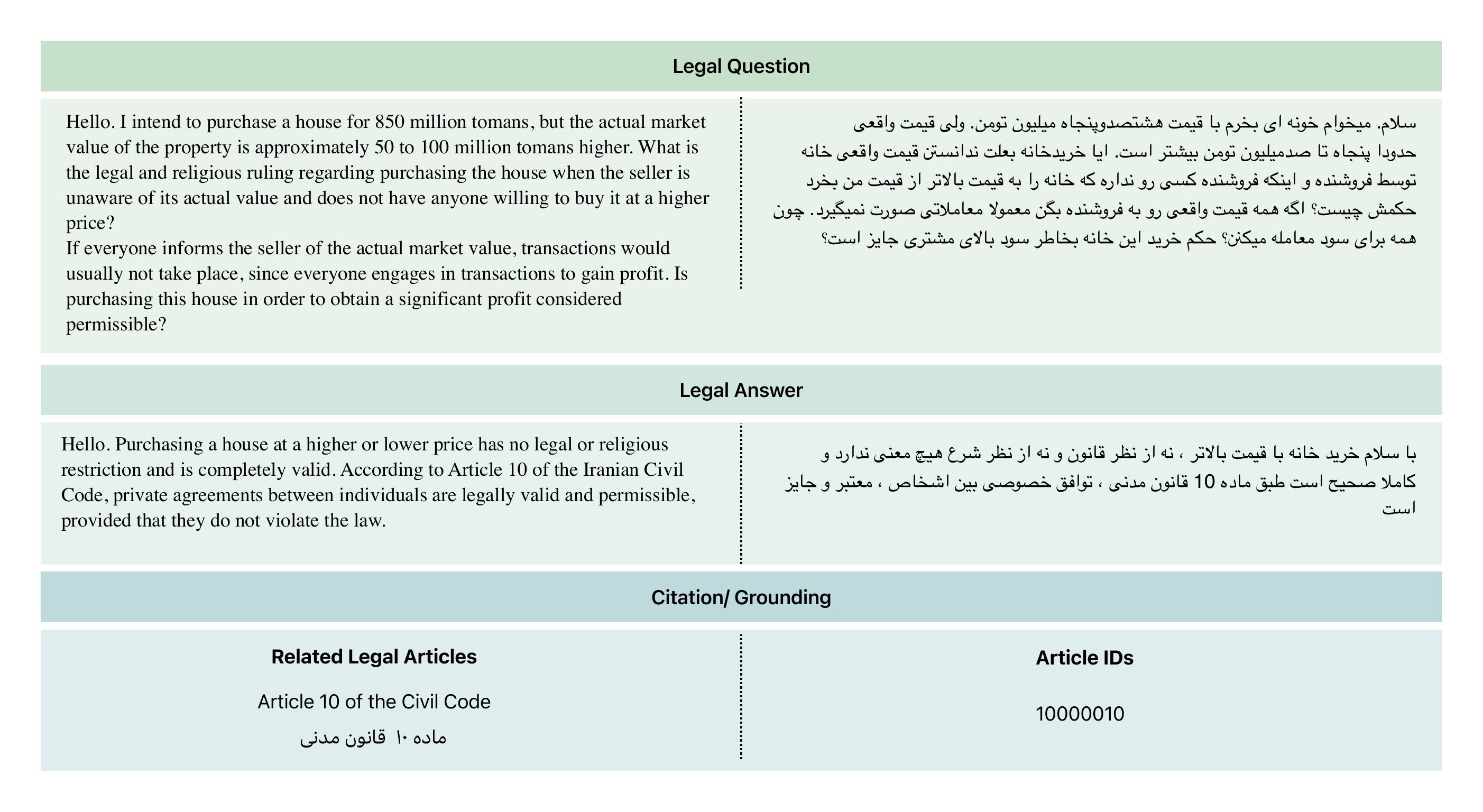}
        \caption{Example instance 2 from TUSI-QA.}
        \label{fig:qa_example_3}
    \end{subfigure}

    \caption{
    Examples of citation-grounded legal question answering instances in TUSI-QA.
    Each instance contains a legal question, the corresponding answer, and the
    referenced legal article providing the supporting citation.
    }
    \label{fig:citation_grounded_qa_examples}
\end{figure}

Figure~\ref{fig:citation_grounded_qa_examples} provides examples of
citation-grounded question answering instances. Each example includes a legal
question, the generated answer, and the supporting legal citation, allowing
evaluation of both answer correctness and the reliability of cited sources.

\begin{figure}[ht]
    \centering

    \begin{subfigure}[b]{0.95\linewidth}
        \centering
        \includegraphics[width=\linewidth]{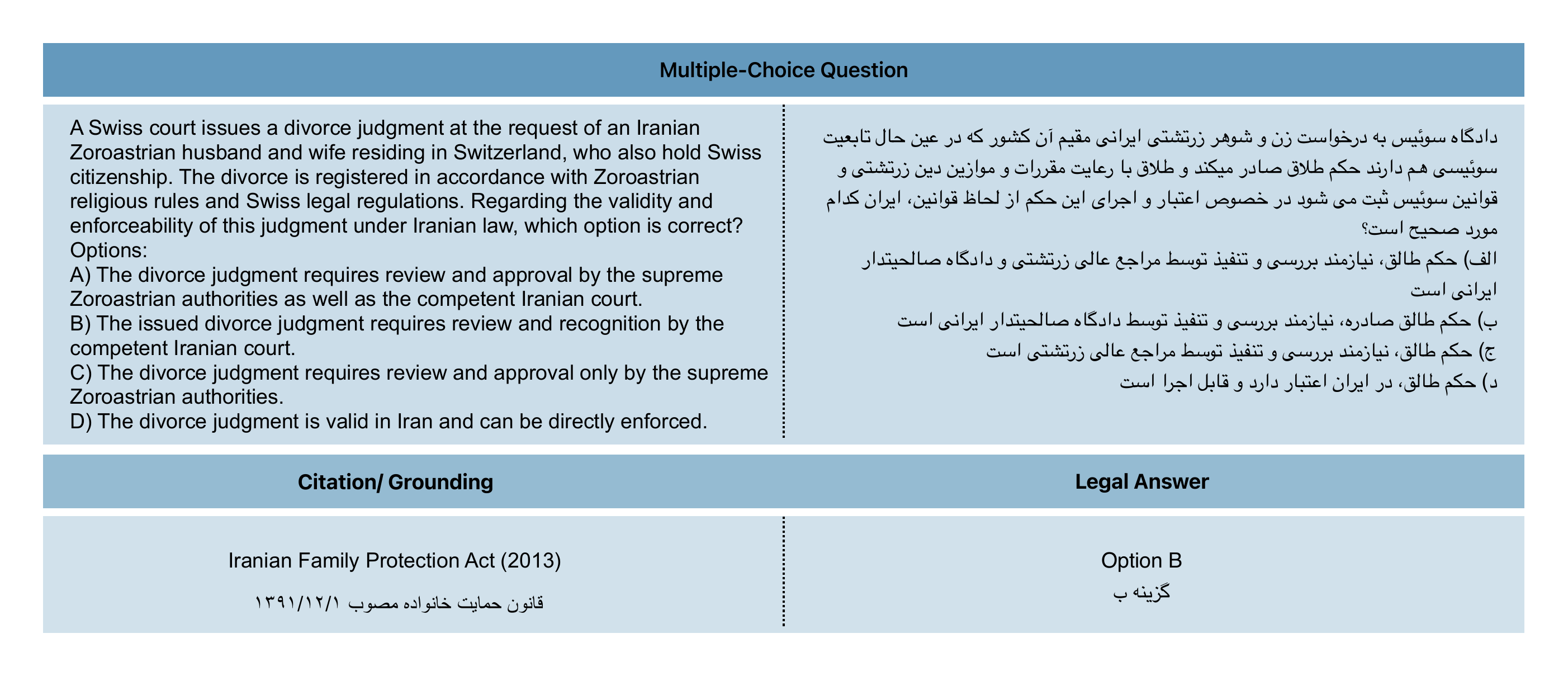}
        \caption{
        Example of a TUSI-Bar multiple-choice question regarding
        the recognition of a foreign divorce judgment and its related
        legal reference.
        }
        \label{fig:bar_exam_example_1}
    \end{subfigure}

    \vspace{0.5cm}

    \begin{subfigure}[b]{0.95\linewidth}
        \centering
        \includegraphics[width=\linewidth]{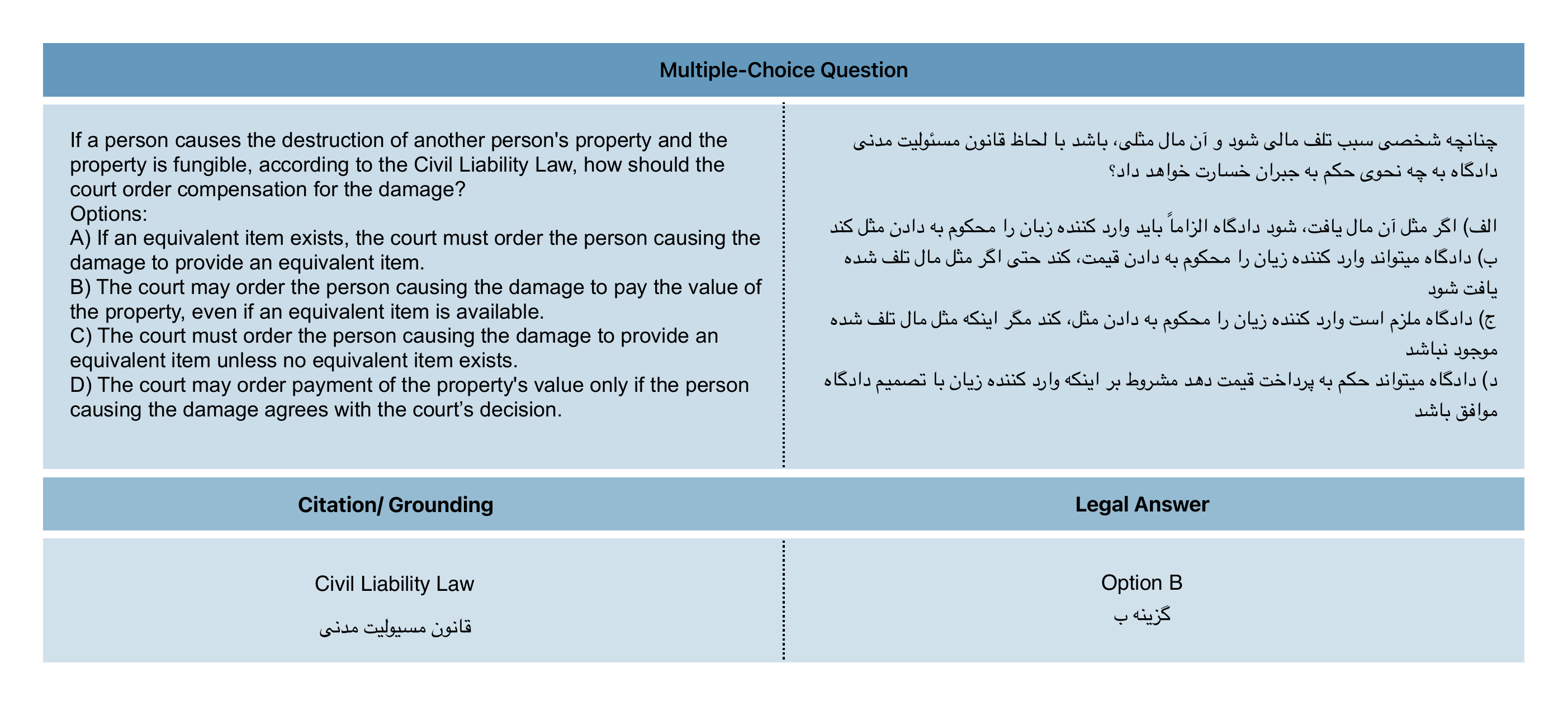}
        \caption{
        Example of a TUSI-Bar multiple-choice question regarding
        compensation for destroyed fungible property.
        }
        \label{fig:bar_exam_example_2}
    \end{subfigure}

    \caption{
    Illustrative examples from TUSI-Bar.
    Each example consists of a multiple-choice legal question, four candidate
    answers, the ground-truth answer, and the associated legal source for
    citation-based legal reasoning.
    }
    \label{fig:bar_exam_dataset_examples}

\end{figure}

Finally, Figure~\ref{fig:bar_exam_dataset_examples} illustrates examples from
TUSI-Bar. Each instance contains a multiple-choice legal reasoning question,
candidate answers, the correct option, and the corresponding legal reference.
These examples highlight the benchmark's focus on evaluating structured legal
reasoning and knowledge application in an exam-style setting.

\end{document}